\documentclass{article}

\makeatletter
\def\input@path{{styles/}}
\makeatother

\usepackage[preprint]{colm2026_conference}
\usepackage{fontspec}

\renewcommand{\encodingdefault}{T1}
\normalfont

\usepackage{microtype}
\usepackage{graphicx}
\usepackage{trimclip}
\usepackage{xcolor}
\usepackage{booktabs}
\usepackage{array}
\usepackage{colortbl}
\usepackage{float}
\usepackage{tikz}
\usepackage{tcolorbox}
\usepackage{pgfplots}
\pgfplotsset{compat=1.18}
\usepgfplotslibrary{groupplots}
\usepackage{hyperref}
\usepackage{url}

\definecolor{abyss}{HTML}{121D36}
\definecolor{polarnight}{HTML}{1A2947}
\definecolor{nebula}{HTML}{2B3F66}
\definecolor{steeltrail}{HTML}{6D87BD}
\definecolor{skytrail}{HTML}{8FA8D8}
\definecolor{starlight}{HTML}{DFE7F5}
\definecolor{warmstar}{HTML}{E8D9C4}
\definecolor{allsparkwordmark}{HTML}{16233F}
\definecolor{allsparkspark}{HTML}{4A659C}
\definecolor{electricblue}{HTML}{3866FF}
\definecolor{covercream}{HTML}{EEF3FA}
\definecolor{coveraccent}{HTML}{3866FF}

\colorlet{pevekpurple}{skytrail}
\colorlet{bargray}{steeltrail}
\colorlet{barlgray}{starlight}

\newfontfamily\outfit[
    Path=assets/fonts/,
    UprightFont=Outfit-Regular.ttf,
    BoldFont=Outfit-SemiBold.ttf
]{Outfit}

\hypersetup{
    colorlinks=true,
    linkcolor=electricblue,
    citecolor=electricblue,
    urlcolor=coveraccent,
    filecolor=electricblue
}

\setcitestyle{numbers,square,comma,sort&compress}

\usepackage{amsmath,amsfonts,bm}

\def\eqref#1{equation~\ref{#1}}

\def\1{\bm{1}}

\DeclareMathAlphabet{\mathsfit}{\encodingdefault}{\sfdefault}{m}{sl}
\SetMathAlphabet{\mathsfit}{bold}{\encodingdefault}{\sfdefault}{bx}{n}

\usepackage{subcaption}

\usepackage{placeins}

\usepackage{amsmath}
\usepackage{amssymb}
\usepackage{mathtools}
\usepackage{amsthm}
\usepackage{enumitem}

\definecolor{rossGroupTint}{HTML}{EAF2F8}
\definecolor{rossRowTint}{HTML}{EAF2F8}
\definecolor{rossTableDivider}{HTML}{9AAFC1}
\definecolor{rossDeltaGain}{HTML}{187B69}
\definecolor{rossDeltaLoss}{HTML}{AE5260}

\usepackage{multicol}
\usepackage{multirow}
\usepackage{makecell}
\usepackage{comment}
\usepackage{wrapfig}
\usepackage{needspace}
\usepackage{bbding}
\usepackage{pifont}

\usepackage{tablefootnote}
\usepackage[normalem]{ulem}

\usepackage{algorithm}
\usepackage{algpseudocode}

\tcbuselibrary{skins,breakable}

\definecolor{MyDarkBlue}{rgb}{0,0.5,1}
\definecolor{MyDarkGreen}{rgb}{0.02,0.6,0.02}
\definecolor{MyDarkRed}{rgb}{0.8,0.02,0.02}
\definecolor{MyDarkOrange}{rgb}{0.40,0.2,0.02}
\definecolor{MyPurple}{RGB}{111,0,255}
\definecolor{MyRed}{rgb}{1.0,0.0,0.0}
\definecolor{MyGold}{rgb}{0.75,0.6,0.12}
\definecolor{MyDarkgray}{rgb}{0.66, 0.66, 0.66}

\usepackage{fontawesome5}

\newcommand{\reporttitle}{ROSS: Relearning from Self-Generated \underline{Ro}llouts through \underline{S}elective \underline{S}upervision}
\title{\reporttitle}
\author{AllSpark Team}
\begin{document}

\fancyhead{}
\renewcommand{\headrulewidth}{0pt}
\color{abyss}
\thispagestyle{empty}

\vspace*{-0.44in}

\begin{tcolorbox}[
    width=\linewidth,
    colback=covercream,
    colframe=covercream,
    boxrule=0pt,
    arc=14pt,
    outer arc=14pt,
    boxsep=0pt,
    left=20pt,
    right=20pt,
    top=13pt,
    bottom=11pt
]

{\outfit\fontsize{21.5}{25.5}\selectfont\bfseries\centering
\textcolor{coveraccent}{ROSS:} Relearning from Self-Generated\\[0.25em]
\underline{Ro}llouts through \underline{S}elective \underline{S}upervision\par}

\vspace{1.45em}

{\bfseries\centering AllSpark Team\par}

\vspace{0.75em}

\begingroup
\normalfont
\setlength{\parindent}{0pt}
\setlength{\parskip}{0pt}
\renewenvironment{abstract}{\ignorespaces}{\par}
\begin{abstract}
Large language model post-training generates self-generated rollouts through reinforcement learning and on-policy distillation, yet this experience is often treated as stale once the policy advances. 
Historical rollouts can remain compatible with a later policy while preserving behaviors that the policy no longer expresses reliably. However, they may also contain mistakes, abandoned attempts, and redundant actions that should not be imitated, motivating finer-grained selective supervision.
We introduce \emph{\textbf{ROSS}} (Relearning from Self-Generated \underline{\textbf{Ro}}llouts through \underline{\textbf{S}}elective \underline{\textbf{S}}upervision), which preserves the full historical trajectory as context while applying loss only to selected model-generated continuations. Across domain-specific reinforcement learning, multi-teacher on-policy distillation, and agentic reinforcement learning, ROSS consistently improves upstream checkpoints and outperforms baselines across mathematics, code generation, instruction following, and software engineering. On Qwen3.6-35B-A3B, ROSS improves the six-benchmark MOPD average from 58.40\% to 62.20\% and SWE-bench Verified from 64.20\% to 68.40\%. These results show that self-rollout training leaves behind reusable behavioral experience that can yield further gains through offline supervised fine-tuning (SFT), without additional policy rollouts.
\end{abstract}
\par
\endgroup

\vspace{0.65em}

\noindent
\begin{minipage}[b]{0.63\linewidth}
    \outfit\fontsize{8.4}{10.2}\selectfont
    \textbf{Date:} September 28, 2026\\
    \textbf{Resources:} \href{https://github.com/AllSpark-Research/ROSS}{GitHub}
\end{minipage}%
\hfill
\begin{minipage}[b]{0.33\linewidth}
    \raggedleft
    \raisebox{-0.30em}{%
        \includegraphics[height=16pt]{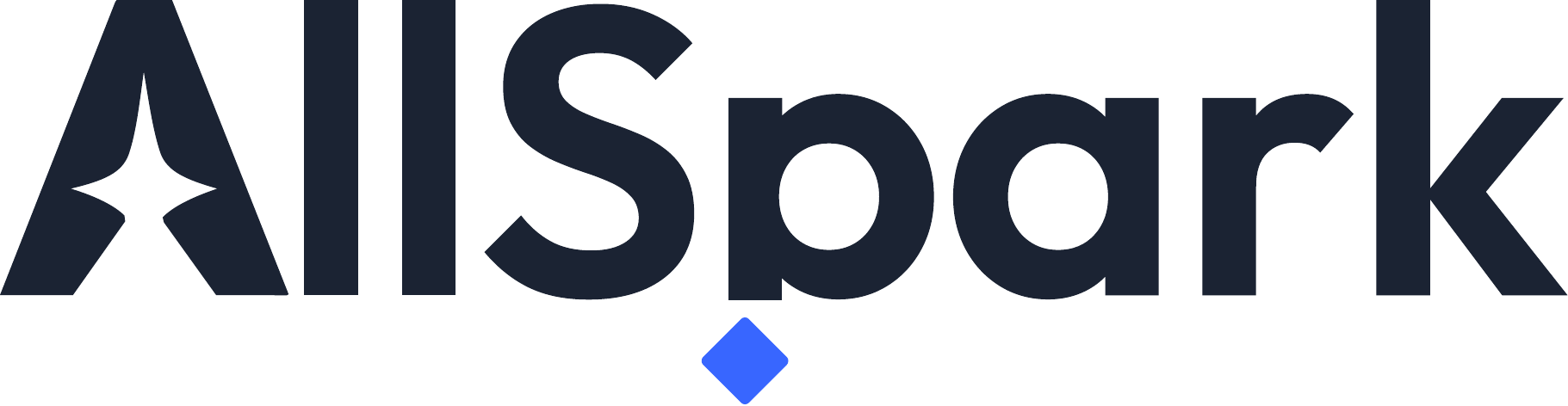}%
    }%
\end{minipage}

\end{tcolorbox}

\vspace{0.1em}


\begin{figure}[h]
    \centering

    \begin{subfigure}[b]{0.44\linewidth}
        \centering
        \includegraphics[width=\linewidth]
        {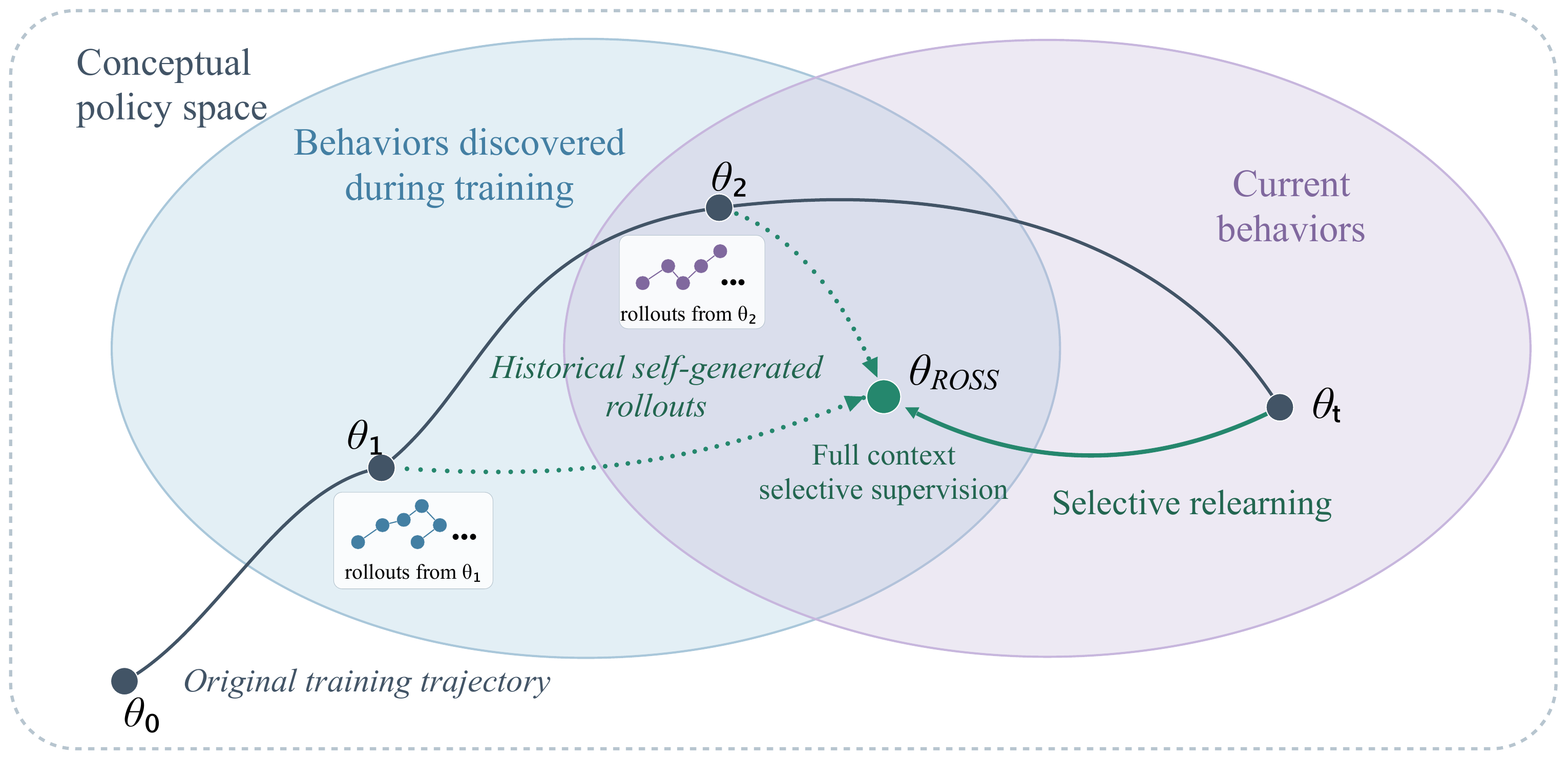}
        \caption{Historical experience preserves discovered behaviors.}
        \label{fig:policy-space}
    \end{subfigure}
    \hfill
    \begin{subfigure}[b]{0.54\linewidth}
        \centering
        \includegraphics[width=\linewidth]
        {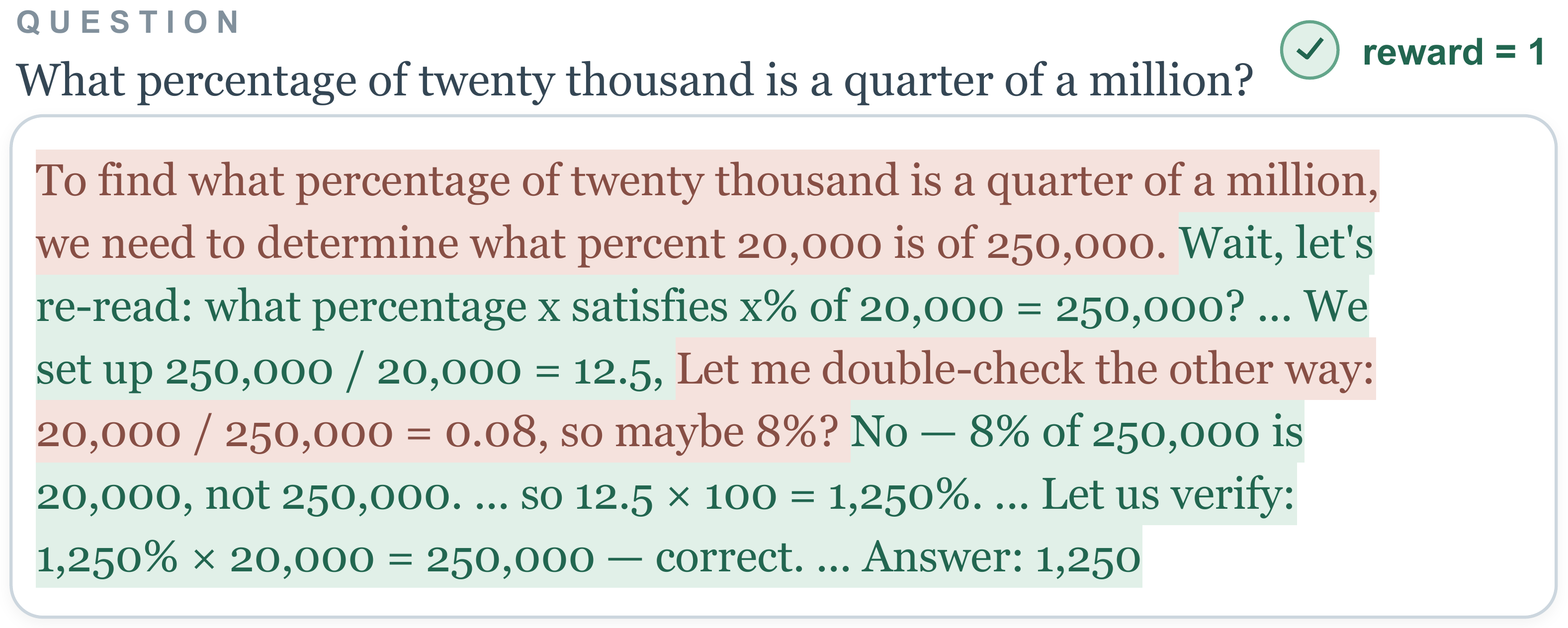}
        \caption{Successful rollouts contain both useful and uninformative segments.}
        \label{fig:selective-supervision}
    \end{subfigure}
    \caption{\textbf{Relearning from self-generated historical rollouts.}
    \textbf{(a)} As the policy evolves, some useful behaviors may become
    underrepresented in its current rollout distribution.
    \textbf{(b)} Even when a rollout reaches a successful outcome, it may include
    redundant reasoning, or unnecessary detours, making full-trajectory replay
    suboptimal.}
    \label{fig:intro}
    \vspace{-5pt}
\end{figure}

\section{Introduction}

Large language models (LLMs)~\citep{guo2025deepseek,lambert2024tulu,yuan2023scaling,yue2025does} are increasingly trained from experience generated by the models themselves. Reinforcement learning (RL)~\citep{yue2025does,guo2025deepseek,hou2026single,zheng2025group} repeatedly samples rollouts from the current policy and updates it based on the resulting feedback, while on-policy distillation (OPD)~\citep{agarwal2024policy, gu2024minillm,lu2025onpolicydistillation,yang2025qwen3,xiao2026mimo,team2026kimi} collects student trajectories and applies teacher supervision to the states they visit. Over training, these procedures produce not only a sequence of model checkpoints, but also a growing record of self-generated experience, including successful reasoning paths, alternative strategies, recoveries, and other behaviors discovered along the way. Yet as training advances, historical rollouts are often treated as stale and discarded. This raises a question: \textit{can a later checkpoint continue to learn from what the model discovered earlier?}

This possibility stems from how the policy evolves during training. As the rollout distribution shifts with the policy, behaviors discovered at earlier stages need not remain reliably expressed at later checkpoints. As illustrated in Figure~\ref{fig:intro}(a), some valid behaviors may become underrepresented even though they remain useful~\citep{zhang2026on,wang2025octothinker,yuan2023scaling,dong2023raft}. Historical rollouts can therefore retain behaviors that the current policy no longer reliably recovers, providing complementary training signal beyond its current rollout distribution~\citep{zheng2026swe,xiong2024watch,slinko2026step}. Yet this complementarity is not uniform across a trajectory: as Figure~\ref{fig:intro}(b) illustrates, even successful rollouts may interleave useful reasoning with mistakes, redundant steps, and unnecessary detours. Outcome-level success alone therefore does not guarantee that every intermediate step provides desirable supervision~\citep{wang2025octothinker,lightman2024let}. The challenge is thus to identify not only which historical rollouts remain valuable, but also which parts of those rollouts are worth relearning.

We propose \emph{\textbf{ROSS}}, \textit{Relearning from Self-Generated \underline{Ro}llouts through \underline{S}elective \underline{S}upervision}, to address these two levels of selection. ROSS identifies valuable historical rollouts and selectively supervises useful segments within successful trajectories while preserving the complete trajectory as context. In this way, the current policy can consolidate previously discovered behaviors without indiscriminately imitating entire trajectories or reinforcing mistakes and unnecessary steps. 
We evaluate ROSS across domain-specific RL, multi-teacher on-policy distillation (MOPD), and agentic RL, spanning reasoning, coding, instruction following, and software engineering. Across these settings, ROSS consistently improves the corresponding upstream checkpoints, including a gain from 58.40\% to 62.20\% on the six-benchmark MOPD average and from 64.20\% to 68.40\% on SWE-bench Verified. These results show that self-rollout training leaves behind not only a stronger policy, but also reusable experience that can be selectively consolidated by later checkpoints.

Our contributions are as follows:
\begin{itemize}[leftmargin=*]
\vspace{-4pt}
\item We establish historical self-generated rollouts as a reusable source of training experience, showing that they can retain useful behaviors under an evolving policy.
\vspace{-2pt}
\item We introduce \emph{\textbf{ROSS}}, which selectively consolidates historical experience by supervising informative segments while preserving full-trajectory context.
\vspace{-2pt}
\item We validate ROSS as an additional offline SFT stage after domain-specific RL, MOPD, and agentic RL, with further gains in math, coding, instruction following, and long-horizon agentic tasks.
\vspace{-4pt}
\end{itemize}

\section{Preliminary Analysis}

\begin{figure}[t]
\centering
\includegraphics[width=\linewidth]{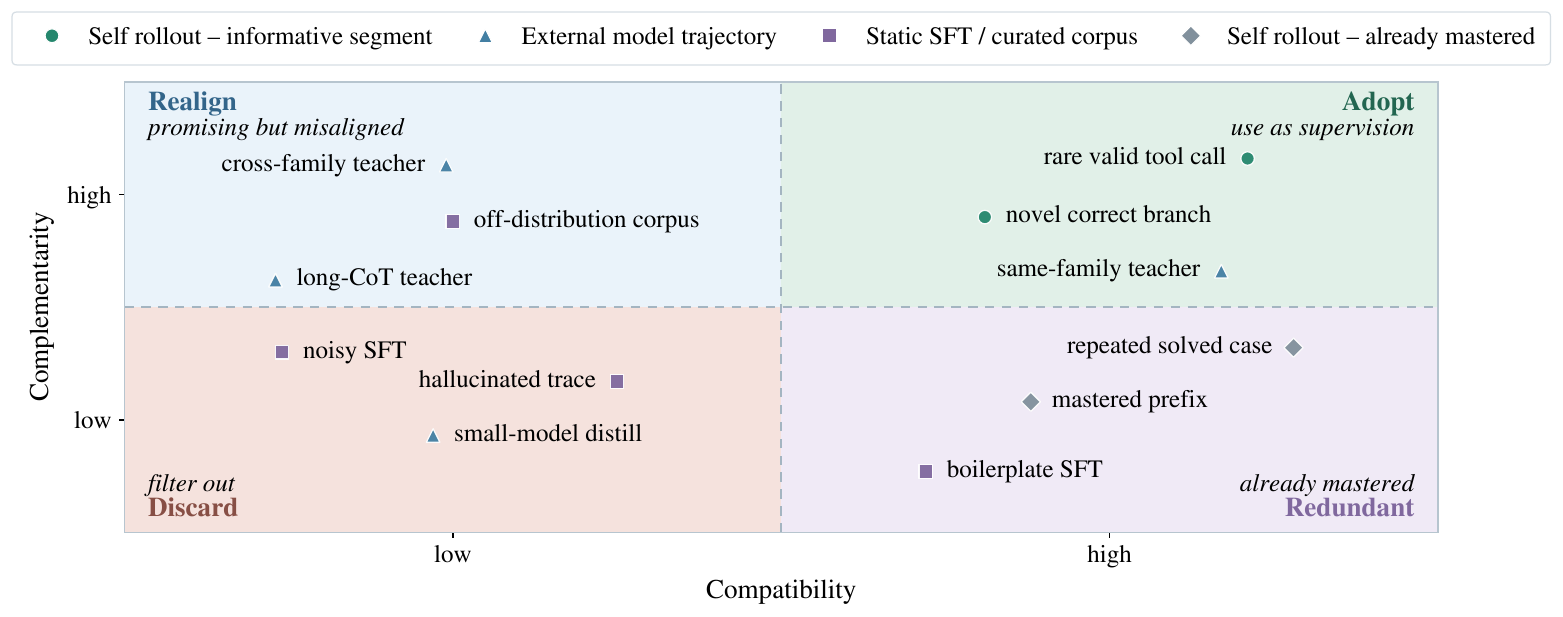}
\vspace{-15pt}
\caption{\textbf{When is historical experience worth revisiting?} Its value depends on \emph{compatibility} with the current policy and \emph{complementarity} to its behavioral coverage.}
\label{fig:preliminary}
\vspace{-15pt}
\end{figure}

To examine when historical experience is worth revisiting, we consider two dimensions: \emph{compatibility} and \emph{complementarity}. Compatibility concerns how well historical outputs align with the current policy's output distribution, while complementarity concerns whether historical experience contains useful behaviors that the current policy does not reliably reproduce. As illustrated in Figure~\ref{fig:preliminary}, these dimensions help distinguish experience worth relearning from behaviors that are poorly aligned with the current policy or already reliably expressed. We examine compatibility through response-level predictability in Section~\ref{sec:compatibility}, and complementarity through success reliability on historically solved problems and a comparison of earlier and later rollouts in Section~\ref{sec:complementarity}.

\begin{figure}[t]
\centering
\includegraphics[width=0.33\linewidth]{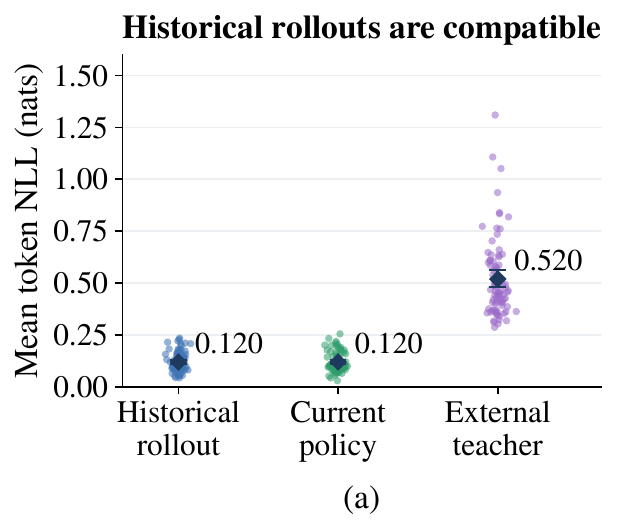}
\includegraphics[width=0.66\linewidth]{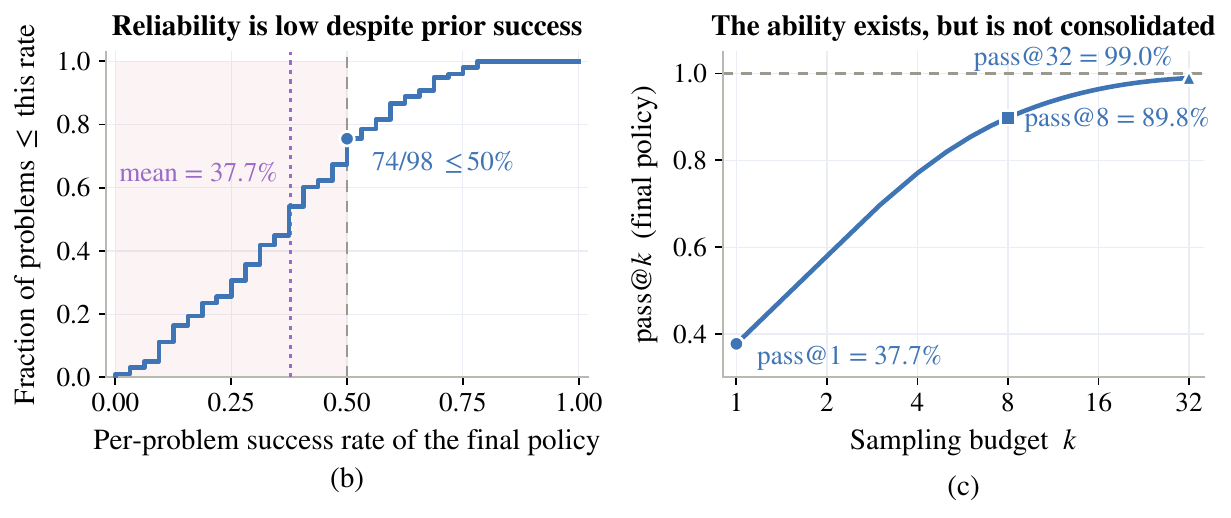}
\vspace{-15pt}
\caption{\textbf{Historical rollouts remain compatible with the current policy, while success on historically solved problems remains unreliable.} \textbf{(a)} Historical rollouts have mean token NLL under the current policy comparable to its own responses and lower than external-teacher responses. \textbf{(b)} On historically solved problems, the current policy achieves a mean $\mathrm{pass@1}$ of only $37.7\%$, indicating low single-attempt reliability. \textbf{(c)} Increasing the sampling budget raises success to $\mathrm{pass@32}=99.0\%$, showing that successful solutions remain attainable but are not reliably generated in a single attempt.}
\vspace{-15pt}
\label{fig:compatibility_consolidation}
\end{figure}

\subsection{Compatibility Analysis}
\label{sec:compatibility}
\vspace{-6pt}

\paragraph{Compatibility provides a basis for relearning that builds on existing capabilities.}
After the original training stage, we seek to consolidate useful behaviors through historical experience while retaining the capabilities already acquired. Supervision targets aligned with the current policy may allow relearning to build on the model's existing behavioral patterns, reducing the additional adaptation needed to fit substantially different outputs and potentially limiting interference with existing capabilities. This motivates us to ask whether historical rollouts generated by earlier checkpoints remain compatible with the current policy.

We select 100 mathematical problems and collect one response per problem from three sources: historical rollouts from earlier checkpoints, rollouts from the current policy, and an external teacher, GLM-5.2~\citep{zeng2026glm}. We compute the mean token-level negative log-likelihood (NLL) of each response under the current policy as a proxy for response-level compatibility. Lower NLL indicates that the current policy assigns higher probabilities, on average in log space, to the response tokens conditioned on their original contexts.

As shown in Figure~\ref{fig:compatibility_consolidation}(a), historical rollouts have substantially lower NLL than external-teacher responses and remain close to current-policy rollouts. On the mathematical problems examined, these findings support historical rollouts as a policy-compatible source of supervision for subsequent relearning, even though they are no longer strictly on-policy~\citep{slinko2026step,yuan2023scaling}.

\subsection{Complementarity Analysis}
\label{sec:complementarity}
\vspace{-4pt}

\paragraph{Historical success remains insufficiently consolidated.}
Compatibility alone does not establish additional learning value: historical outputs may already be reliably generated by the current policy. We therefore examine whether the current policy consistently succeeds on problems that were solved in historical rollouts.

We select the $100$ most difficult training problems and collect $32$ current-policy rollouts for each, retaining the $98$ problems for which at least one historical rollout was successful. As shown in Figure~\ref{fig:compatibility_consolidation}(c), the current policy solves $97$ of these problems within $32$ attempts, achieving $\mathrm{pass@32}=99.0\%$. Successful solutions to these problems therefore remain attainable under the current policy. However, as shown in Figure~\ref{fig:compatibility_consolidation}(b), the mean $\mathrm{pass@1}$ is only $37.7\%$, and $74$ of the $98$ problems have a per-problem success rate of at most $50\%$. These results reveal a gap between generating a successful solution and doing so reliably. Historical rollouts preserve successful demonstrations for these unreliably solved problems; together with the compatibility results in Section~\ref{sec:compatibility}, they suggest a source of policy-compatible supervision for consolidating task performance that remains unreliable.

\vspace{-6pt}

\paragraph{Historical experience retains useful behaviors that later rollouts may fail to reproduce.}
Appendix~\ref{app:segment_complementarity_case} illustrates this complementarity with a concrete example. Both an earlier historical rollout and a later rollout arrive at the same incorrect intermediate count, $2^{10}=1024$. The historical rollout then identifies an overlooked constraint and corrects the count to $1007$, whereas the later rollout stops at $1024$. The historical trajectory thus records a useful correction that is absent from the later response, providing a concrete supervision target for relearning. This example illustrates how historical experience preserves useful behaviors that later rollouts may fail to reproduce.

These analyses suggest that historical rollouts can remain compatible with the current policy while preserving useful behaviors that it does not reliably reproduce. This motivates ROSS to revisit experience accumulated during training and select valuable supervision for relearning.

\section{Method}
\label{sec:method}

\begin{figure}[t]
    \centering
    \includegraphics[width=0.98\linewidth]{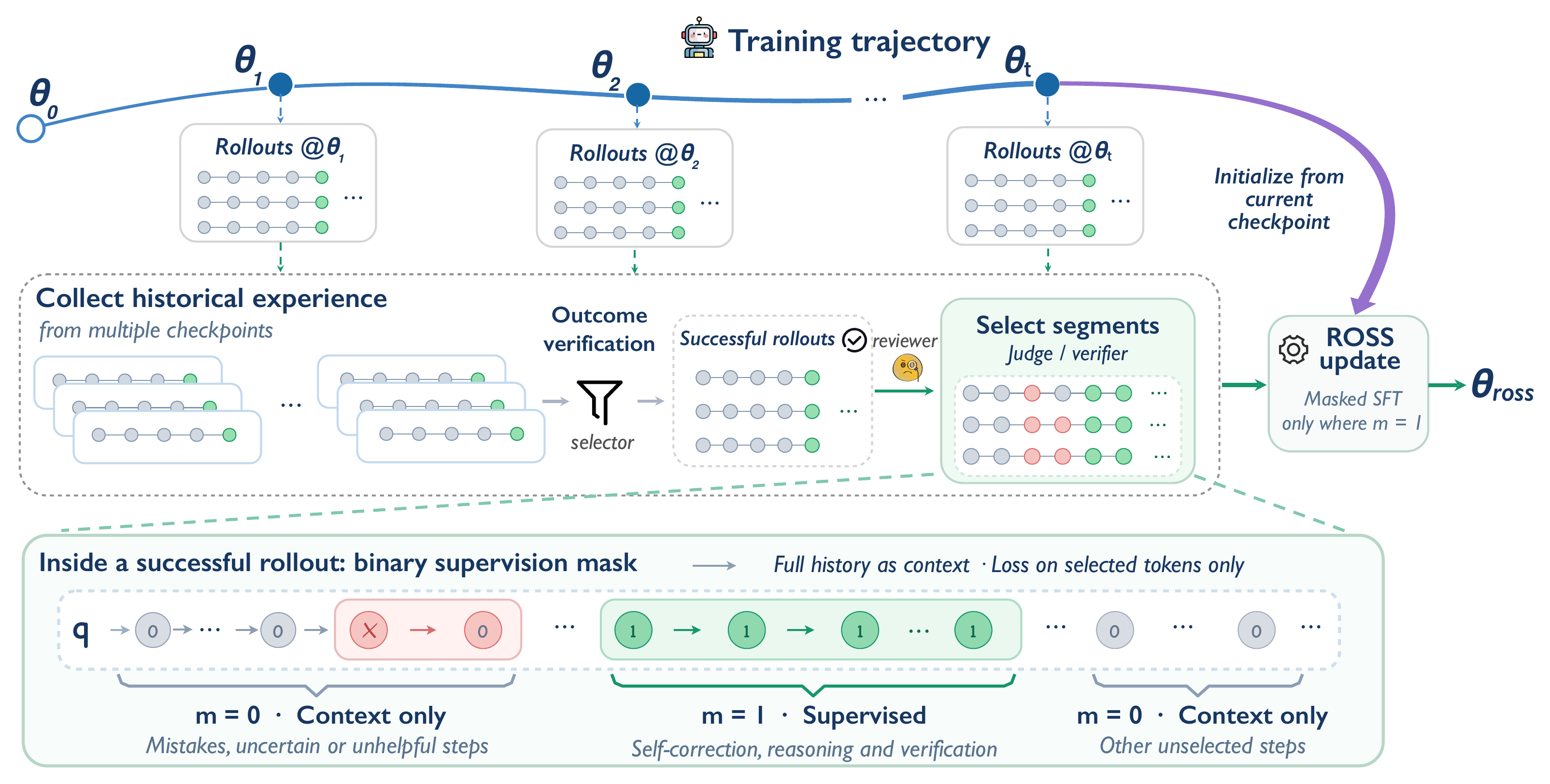}
    \caption{\textbf{Overview of ROSS.} Starting from the final checkpoint, ROSS filters saved rollouts by outcome verification, reviews the positive trajectories, and applies imitation loss only to selected model-generated tokens while preserving the full history as context.}
    \label{fig:method_overview}
    \vspace{-4pt}
\end{figure}

ROSS relearns from historical self-generated rollouts through selective supervision. Starting from the final checkpoint of the original training run, an outcome verifier first identifies positive trajectories. A staged annotation procedure then uses an LLM reviewer to assess each candidate. If a reliable imitation target can be extracted, the reviewer identifies the model-generated tokens that should receive imitation loss. Figure~\ref{fig:method_overview} summarizes the procedure.

\subsection{Problem Formulation}
\label{sec:method_formulation}

Consider a self-rollout training procedure that produces a sequence of policy checkpoints
$\{\pi_{\theta_k}\}_{k=0}^{T}$ and rollout batches $\{\mathcal{D}_k\}_{k=0}^{T-1}$, where
$\tau\sim\pi_{\theta_k}$ for every $\tau\in\mathcal{D}_k$. The original training may use reinforcement learning, on-policy distillation, or agentic reinforcement learning; ROSS requires only the resulting checkpoints and saved rollouts. We define the historical experience pool as
\begin{equation}
    \mathcal{H}_{<T}=\bigcup_{k<T}\mathcal{D}_k,
\end{equation}
and initialize relearning from the final checkpoint $\pi_{\theta_T}$. Although $\mathcal{H}_{<T}$ is no longer strictly on-policy for $\pi_{\theta_T}$, its rollouts were generated by earlier checkpoints from the same training run and may contain behaviors that $\pi_{\theta_T}$ does not reliably express.

We represent each rollout as $\tau=(z_1,\ldots,z_L)$, which may encode a single-turn exchange or an agentic interaction history. Let $a_t\in\{0,1\}$ indicate whether $z_t$ is a policy-generated token eligible for imitation; prompts, environment observations, and other exogenous context have $a_t=0$. ROSS applies a trajectory-level selector $g(\tau)\in\{0,1\}$ and, for retained rollouts, a token-level supervision mask $m_t\in\{0,1\}$. Staged LLM review followed by deterministic boundary and token-alignment checks produces a mask satisfying $m_t\leq a_t$.

\subsection{Selective Supervision over Historical Rollouts}
\label{sec:method_selection}

\paragraph{Trajectory-level selection.}
Let $v(\tau)$ denote the task outcome verifier, such as an exact-answer check, executable checker, unit-test suite, or environment success signal. It supplies trajectory-level feedback only. We retain
\begin{equation}
    \mathcal{H}^{+}=\{\tau\in\mathcal{H}_{<T}:g(\tau)=1\},
    \qquad g(\tau)=\mathbf{1}[v(\tau)=1].
\end{equation}

\paragraph{Within-trajectory selection.}
For each $\tau\in\mathcal{H}^{+}$, a staged LLM-based annotation procedure $A$ uses an LLM reviewer to examine the task, complete trajectory, and verifier or environment evidence. It returns an audit status $q_\tau$ and disjoint, ordered intervals
$\mathcal{K}_{\tau}=\{[b_j,e_j)\}_{j=1}^{J_\tau}$ that contain locally correct, self-contained, and behaviorally useful model outputs. We convert these intervals into a token mask
\begin{equation}
    m_t(\tau)=a_t\cdot
    \mathbf{1}\!\left[t\in\bigcup_{[b,e)\in\mathcal{K}_\tau}[b,e)\right].
    \label{eq:ross_mask}
\end{equation}
Proposed targets are independently audited. A positive verifier outcome alone does not guarantee a valid imitation target: a response may reach the correct final answer despite an invalid derivation. A candidate is rejected when no valid target can be recovered from the recorded trajectory without retaining a confirmed error or introducing missing reasoning. The resulting training subset is
\[
    \mathcal{H}_{\mathrm{ROSS}}=
    \left\{\tau\in\mathcal{H}^{+}:
    q_\tau=\textsc{accept},\;
    \mathcal{K}_\tau\neq\varnothing,\;
    \text{\textsc{Validate}}(\tau,m(\tau))=1
    \right\}.
\]
Here, \textsc{Validate} denotes the deterministic checks in Algorithm~\ref{alg:ross} that ensure source consistency and token alignment. Single-turn responses use positive-span extraction, whereas agentic histories use defect localization; both are ultimately compiled into supervised token spans and the same token-level mask. Appendix~\ref{app:annotation} details the annotation protocols and core reviewer prompts.

\paragraph{Full context, selective targets.}
Setting $m_t=0$ removes a token from the loss, not from the sequence. A selected token $z_t$ is therefore predicted from the complete original prefix $z_{<t}$, including earlier mistakes, abandoned attempts, and environment feedback, preserving the state in which the continuation originally occurred.

\subsection{ROSS Training Objective}
\label{sec:method_objective}

Starting from $\theta_T$, ROSS performs masked teacher-forcing on the selected historical rollouts. Its objective is
\begin{equation}
    \mathcal{L}_{\mathrm{ROSS}}(\theta)
    =-
    \frac{
        \sum_{\tau\in\mathcal{H}_{\mathrm{ROSS}}}
        \sum_{t=1}^{|\tau|}m_t(\tau)
        \log\pi_{\theta}(z_t\mid z_{<t})
    }{
        \sum_{\tau\in\mathcal{H}_{\mathrm{ROSS}}}
        \sum_{t=1}^{|\tau|}m_t(\tau)
    }.
    \label{eq:ross_objective}
\end{equation}
Setting $m_t=a_t$ for every $\tau\in\mathcal{H}_{\mathrm{ROSS}}$ recovers ROSS w/o mask, which uses the same post-annotation subset as ROSS but supervises all eligible policy-generated tokens. Positive-Rollout SFT instead supervises all eligible policy-generated tokens in all verifier-positive trajectories in $\mathcal{H}^{+}$. Appendix~\ref{app:algorithm} states the full procedure as Algorithm~\ref{alg:ross}.

\section{Experiments}
\label{sec:experiments}

\subsection{Experimental Setup}

\paragraph{Models and training.}
We study two forms of self-generated experience with Qwen3.6-35B-A3B: single-turn rollouts and agentic trajectories. \textit{For single-turn experiments}, we reuse historical rollouts from independent mathematics and code RL runs, as well as multi-teacher on-policy distillation (MOPD) spanning mathematics, code, and instruction following. ROSS starts from the final checkpoint and applies masked SFT to these rollouts. All single-turn experiments use a no-thinking configuration and report results after three SFT epochs, with optimization settings matched within each setting. Rollout review and mask annotation use GLM-5.2 in high-thinking mode as the LLM judge, which reviews the saved rollouts without generating replacement targets. \textit{For agentic scenarios}, we train the model with agentic RL on OpenSWE tasks from daVinci-Env~\citep{fu2026davinci}, using a Codex CLI agent~\citep{openai2025codex} to interact with containerized repositories through Harbor~\citep{harbor2026framework}. The task verifier provides terminal rewards. Both agentic training and relearning retain thinking traces and use a separate long-context schedule. Detailed configurations, rollout windows, and annotation protocols are provided in Appendices~\ref{app:experimental_details} and~\ref{app:annotation}.

\paragraph{Comparisons.}
We compare against four baselines: \emph{(i) Base}, the open-source checkpoint used to initialize the original RL or MOPD training; \emph{(ii) Upstream}, the final checkpoint of the original RL or MOPD run; \emph{(iii) Continued RL/MOPD}, which follows the original training procedure for two additional epochs in Math and Code RL, or for 200 additional steps from the 200-step MOPD checkpoint; and \emph{(iv) Positive-Rollout SFT}, which applies SFT to verifier-positive historical rollouts. 

\paragraph{Evaluation.}
Single-turn evaluation covers mathematics (AIME 2025, AIME 2026, HMMT-November 2025), code generation (LCB Gen, OJBench), and instruction following (IFBench). Agentic coding is evaluated on SWE-bench Verified~\citep{jimenez2024swebench} in thinking mode, with the evaluation scaffold, task environments, tool interfaces, and execution budgets fixed across policies. Details of the evaluation setup are given in Appendix~\ref{app:experimental_details}.

\subsection{Main Results}
\begin{table}[t]
\centering
\caption{\textbf{Main results with Qwen3.6-35B-A3B.} All scores are higher-is-better. (a) Math and Code use independent domain-specific RL runs; Math Avg. and Code Avg. are the corresponding domain means. (b) MOPD spans all three domains, and Avg. is the mean of its six benchmarks. All relearning methods start from the corresponding Upstream checkpoint; bold marks the best result among them, including ties.}
\label{tab:main_results}
\begingroup
\small
\setlength{\tabcolsep}{5pt}
\renewcommand{\arraystretch}{1.15}
\textit{(a) Domain-specific RL}\par\smallskip
\resizebox{\linewidth}{!}{%
\begin{tabular}{@{}l cccc ccc@{}}
\toprule
& \multicolumn{4}{c}{\textsc{Math RL}} & \multicolumn{3}{c}{\textsc{Code RL}} \\
\cmidrule(lr){2-5}\cmidrule(l){6-8}
Method & AIME 25 & AIME 26 & HMMT-Nov. & Math Avg. & LCB Gen & OJBench & Code Avg. \\
\midrule
Base & 71.25 & 76.56 & 70.62 & 72.81 & 57.24 & 22.20 & 39.72 \\
Upstream & 74.84 & 77.50 & 73.23 & 75.19 & 58.95 & 25.22 & 42.09 \\
Continued RL & 74.64 & 78.85 & 72.81 & 75.43 & 57.24 & \textbf{27.37} & 42.31 \\
\midrule
Positive-Rollout SFT & 75.47 & 79.06 & 73.44 & 75.99 & 58.19 & 23.49 & 40.84 \\
\rowcolor{rossRowTint}
\textbf{ROSS (ours)} & \textbf{76.46} & \textbf{79.90} & \textbf{74.58} & \textbf{76.98} & \textbf{61.05} & \textbf{27.37} & \textbf{44.21} \\
\bottomrule
\end{tabular}%
}
\par\medskip
\textit{(b) MOPD}\par\smallskip
\resizebox{\linewidth}{!}{%
\begin{tabular}{@{}l ccc cc c c@{}}
\toprule
& \multicolumn{3}{c}{\textsc{Math}} & \multicolumn{2}{c}{\textsc{Code}} & \textsc{IF} & \\
\cmidrule(lr){2-4}\cmidrule(lr){5-6}\cmidrule(l){7-7}
Method & AIME 25 & AIME 26 & HMMT-Nov. & LCB Gen & OJBench & IFBench & Avg. \\
\midrule
Base & 71.25 & 76.56 & 70.62 & 57.24 & 22.20 & 34.10 & 55.33 \\
Upstream & 72.81 & 77.60 & 71.25 & 57.43 & 26.72 & 44.58 & 58.40 \\
Continued MOPD & 73.39 & 78.23 & 72.60 & 57.71 & 27.59 & 45.45 & 59.16 \\
\midrule
Positive-Rollout SFT & 74.06 & 77.66 & 69.17 & 56.86 & 26.94 & 41.49 & 57.70 \\
\rowcolor{rossRowTint}
\textbf{ROSS (ours)} & \textbf{77.76} & \textbf{80.42} & \textbf{76.77} & \textbf{60.76} & \textbf{29.74} & \textbf{47.76} & \textbf{62.20} \\
\bottomrule
\end{tabular}%
}
\endgroup
\end{table}

We evaluate ROSS on historical rollouts from domain-specific RL and MOPD, comparing it with the upstream checkpoints and baseline training strategies. Table~\ref{tab:main_results} summarizes the results. Across both settings, ROSS consistently achieves the best performance among all relearning methods. After domain-specific RL, ROSS achieves the highest average on both Math and Code, improving from 75.19 to 76.98 and from 42.09 to 44.21, respectively. After MOPD, ROSS improves the six-benchmark average from 58.40 to 62.20, outperforming Continued MOPD and Positive-Rollout SFT. These results show that historical rollouts retain useful learning signal beyond their original training stage. While continued RL/MOPD further explores the policy's behavioral frontier, ROSS complements this process by consolidating useful behaviors accumulated throughout training that may remain unreliably expressed by the current checkpoint.

We further examine whether ROSS introduces degradation outside the domain targeted by domain-specific RL. Table~\ref{tab:ood_retention} in Appendix~\ref{app:ood_retention} compares each ROSS-retrained checkpoint with its corresponding upstream RL checkpoint on mathematics, code generation, instruction following, and agentic tool use. ROSS preserves the gains of the domain-specific checkpoints while improving several off-domain capabilities. After Math RL, it raises the Avg. Code from 41.21 to 44.24 and IFBench from 33.30 to 34.60; after Code RL, it improves Avg. Math from 74.19 to 75.03. ROSS also improves multi-turn agentic tool use, raising the overall BFCL score from 44.12 to 45.62 and from 46.25 to 49.38, respectively. These cross-domain gains are consistent with the near-on-policy compatibility of historical rollouts demonstrated in Section~\ref{sec:compatibility}, suggesting that relearning from them can consolidate useful experience without sacrificing capabilities acquired by the current policy.
\Needspace{20\baselineskip}
\subsection{Relearning from Agentic Trajectories}
\label{sec:agentic_results}

\begin{wraptable}[11]{r}{0.5\linewidth}
\vspace{-6pt}
\centering
\caption{\textbf{Agentic trajectory reuse.} $\Delta$ is relative to Upstream.}
\vspace{-5pt}
\label{tab:agentic_main}
\small
\setlength{\tabcolsep}{5pt}
\begin{tabular}{lc}
\toprule
Method & SWE-bench Verified $\uparrow (\Delta)$ \\
\midrule
Base & 60.80 \\
Upstream & 64.20 \\
Positive-Rollout SFT & 65.20 {\scriptsize \textcolor{rossDeltaGain}{$(+1.00)$}} \\
\rowcolor{rossRowTint}
\textbf{ROSS (ours)} & \textbf{68.40} {\scriptsize \textcolor{rossDeltaGain}{$\mathbf{(+4.20)}$}} \\
\bottomrule
\end{tabular}
\vspace{-5pt}
\end{wraptable}

The results above establish the value of historical experience in single-turn rollouts. We next examine whether this benefit extends to long-horizon agent--environment interaction. We reuse successful OpenSWE trajectories from the agentic RL run and evaluate on SWE-bench Verified, with the training protocol described in Appendix~\ref{app:agentic_setup}.
As shown in Table~\ref{tab:agentic_main}, ROSS improves the resolved-issue rate from 64.20 to 68.40, yielding a 4.20-point gain over the Upstream RL checkpoint. The gain shows that ROSS generalizes beyond single-turn rollouts to long-horizon agentic trajectories, where useful behaviors are embedded in extended interaction histories.

\FloatBarrier

\begin{table}[!t]
\centering
\caption{\textbf{Effect of trajectory filtering and token-level masking.}
Subscripts show percentage-point changes from Upstream; green/red denote gains/losses and bold marks the best score. ROSS w/o mask and ROSS use identical examples and differ only by token-level masking.}
\vspace{-8pt}
\label{tab:decomposition}
\begingroup
\small
\setlength{\tabcolsep}{5pt}
\renewcommand{\arraystretch}{1.2}
\textit{(a) Domain-specific RL}\par\smallskip
{\footnotesize
\begin{tabular}{@{}>{\raggedright\arraybackslash}p{0.20\linewidth}*{5}{>{\centering\arraybackslash}p{0.135\linewidth}}@{}}
\toprule
& \multicolumn{3}{c}{\textsc{Math RL}} & \multicolumn{2}{c}{\textsc{Code RL}} \\
\cmidrule(lr){2-4}\cmidrule(l){5-6}
Method & AIME 25 & AIME 26 & HMMT-Nov. & LCB Gen & OJBench \\
\midrule
Upstream & 74.84 & 77.50 & 73.23 & 58.95 & 25.22 \\
Positive-Rollout SFT & 75.47\textsubscript{\textcolor{rossDeltaGain}{\scriptsize $+0.63$}} & 79.06\textsubscript{\textcolor{rossDeltaGain}{\scriptsize $+1.56$}} & 73.44\textsubscript{\textcolor{rossDeltaGain}{\scriptsize $+0.21$}} & 58.19\textsubscript{\textcolor{rossDeltaLoss}{\scriptsize $-0.76$}} & 23.49\textsubscript{\textcolor{rossDeltaLoss}{\scriptsize $-1.73$}} \\
ROSS w/o mask & \textbf{76.67}\textsubscript{\textcolor{rossDeltaGain}{\scriptsize $+1.83$}} & 79.06\textsubscript{\textcolor{rossDeltaGain}{\scriptsize $+1.56$}} & 73.23\textsubscript{\textcolor{gray}{\scriptsize $+0.00$}} & 59.14\textsubscript{\textcolor{rossDeltaGain}{\scriptsize $+0.19$}} & 26.94\textsubscript{\textcolor{rossDeltaGain}{\scriptsize $+1.72$}} \\
\rowcolor{rossGroupTint}
\textbf{ROSS (ours)} & 76.46\textsubscript{\textcolor{rossDeltaGain}{\scriptsize $+1.62$}} & \textbf{79.90}\textsubscript{\textcolor{rossDeltaGain}{\scriptsize $+2.40$}} & \textbf{74.58}\textsubscript{\textcolor{rossDeltaGain}{\scriptsize $+1.35$}} & \textbf{61.05}\textsubscript{\textcolor{rossDeltaGain}{\scriptsize $+2.10$}} & \textbf{27.37}\textsubscript{\textcolor{rossDeltaGain}{\scriptsize $+2.15$}} \\
\bottomrule
\end{tabular}}
\par\medskip
\textit{(b) Multi-teacher on-policy distillation}\par\smallskip
\resizebox{\linewidth}{!}{%
\begin{tabular}{@{}l@{\hspace{10pt}}ccc cc cc@{}}
\toprule
Method & AIME 25 & AIME 26 & HMMT-Nov. & LCB Gen & OJBench & IFBench & Avg. \\
\midrule
Upstream & 72.81 & 77.60 & 71.25 & 57.43 & 26.72 & 44.58 & 58.40 \\
Positive-Rollout SFT & 74.06\textsubscript{\textcolor{rossDeltaGain}{\scriptsize $+1.25$}} & 77.66\textsubscript{\textcolor{rossDeltaGain}{\scriptsize $+0.06$}} & 69.17\textsubscript{\textcolor{rossDeltaLoss}{\scriptsize $-2.08$}} & 56.86\textsubscript{\textcolor{rossDeltaLoss}{\scriptsize $-0.57$}} & 26.94\textsubscript{\textcolor{rossDeltaGain}{\scriptsize $+0.22$}} & 41.49\textsubscript{\textcolor{rossDeltaLoss}{\scriptsize $-3.09$}} & 57.70\textsubscript{\textcolor{rossDeltaLoss}{\scriptsize $-0.70$}} \\
ROSS w/o mask & 73.28\textsubscript{\textcolor{rossDeltaGain}{\scriptsize $+0.47$}} & 77.55\textsubscript{\textcolor{rossDeltaLoss}{\scriptsize $-0.05$}} & 71.67\textsubscript{\textcolor{rossDeltaGain}{\scriptsize $+0.42$}} & 58.38\textsubscript{\textcolor{rossDeltaGain}{\scriptsize $+0.95$}} & 27.59\textsubscript{\textcolor{rossDeltaGain}{\scriptsize $+0.87$}} & 42.45\textsubscript{\textcolor{rossDeltaLoss}{\scriptsize $-2.13$}} & 58.49\textsubscript{\textcolor{rossDeltaGain}{\scriptsize $+0.09$}} \\
\rowcolor{rossGroupTint}
\textbf{ROSS (ours)} & \textbf{77.76}\textsubscript{\textcolor{rossDeltaGain}{\scriptsize $+4.95$}} & \textbf{80.42}\textsubscript{\textcolor{rossDeltaGain}{\scriptsize $+2.82$}} & \textbf{76.77}\textsubscript{\textcolor{rossDeltaGain}{\scriptsize $+5.52$}} & \textbf{60.76}\textsubscript{\textcolor{rossDeltaGain}{\scriptsize $+3.33$}} & \textbf{29.74}\textsubscript{\textcolor{rossDeltaGain}{\scriptsize $+3.02$}} & \textbf{47.76}\textsubscript{\textcolor{rossDeltaGain}{\scriptsize $+3.18$}} & \textbf{62.20}\textsubscript{\textcolor{rossDeltaGain}{\scriptsize $+3.80$}} \\
\bottomrule
\end{tabular}}
\endgroup
\end{table}

\subsection{What Drives Relearning from Historical Rollouts?}

Having shown that historical rollouts remain useful across both single-turn and agentic settings, we analyze two factors behind these gains: whether token-level masking adds value beyond trajectory-level filtering, and whether relearning depends on inheriting the Upstream parameter updates.
\vspace{-8pt}
\paragraph{The Role of Trajectory Filtering and Selective Supervision}
\label{sec:decomposition}

We disentangle trajectory filtering from token-level selective supervision. Table~\ref{tab:decomposition} compares Positive-Rollout SFT on all verifier-positive rollouts, ROSS w/o mask on the rollouts retained by ROSS with all eligible tokens supervised, and full ROSS. We find that: \textit{(i)} Trajectory filtering already provides a clear benefit. ROSS w/o mask reverses the degradation of Positive-Rollout SFT on Code, improving over Upstream by 0.19 points on LCB Gen and 1.72 points on OJBench, while raising the MOPD average from 57.70 to 58.49, slightly above Upstream checkpoint. \textit{(ii)} Selective supervision accounts for the remaining gains. With the replay set fixed, ROSS further improves over ROSS w/o mask by 3.71 points on MOPD, 1.91 on LCB Gen, 0.43 on OJBench, and 0.66 on Math, showing that selectively supervising informative segments is more effective than imitating all eligible tokens.

\begin{table}[!t]
\centering
\caption{\textbf{ROSS reaches similar scores from different initializations.} Within each setting, the two runs use identical historical trajectories, masks, and SFT configurations. Math is Avg. Math, Code is LCB Gen, and IF is IFBench.}
\vspace{-5pt}
\label{tab:initialization}
\begingroup
\small
\setlength{\tabcolsep}{6pt}
\renewcommand{\arraystretch}{1.08}
\begin{tabular}{@{}lccccc@{}}
\toprule
& Math RL & Code RL & \multicolumn{3}{c}{MOPD} \\
\cmidrule(lr){2-2}\cmidrule(lr){3-3}\cmidrule(l){4-6}
& Math $\uparrow$ & Code $\uparrow$ & Math $\uparrow$ & Code $\uparrow$ & IF $\uparrow$ \\
\midrule
Base & 72.81 & 57.24 & 72.81 & 57.24 & 34.10 \\
Upstream & 75.19 & 58.95 & 73.89 & 57.43 & 44.58 \\
\midrule
\multicolumn{6}{@{}l}{\textit{After ROSS on identical trajectories and masks}} \\
\rowcolor{rossGroupTint}
\textbf{Initialized from Base}
& \textbf{77.09}
& \textbf{61.33}
& \textbf{77.95}
& \textbf{61.24}
& \textbf{47.18} \\
\rowcolor{rossGroupTint}
\textbf{Initialized from Upstream}
& \textbf{76.98}\textsubscript{\textcolor{rossDeltaLoss}{\scriptsize $\mathbf{-0.11}$}}
& \textbf{61.05}\textsubscript{\textcolor{rossDeltaLoss}{\scriptsize $\mathbf{-0.28}$}}
& \textbf{78.32}\textsubscript{\textcolor{rossDeltaGain}{\scriptsize $\mathbf{+0.37}$}}
& \textbf{60.76}\textsubscript{\textcolor{rossDeltaLoss}{\scriptsize $\mathbf{-0.48}$}}
& \textbf{47.76}\textsubscript{\textcolor{rossDeltaGain}{\scriptsize $\mathbf{+0.58}$}} \\
\bottomrule
\end{tabular}
\endgroup
\par\vspace{12pt}
\begingroup
\captionsetup{type=figure}
\centering
\includegraphics[width=\linewidth]{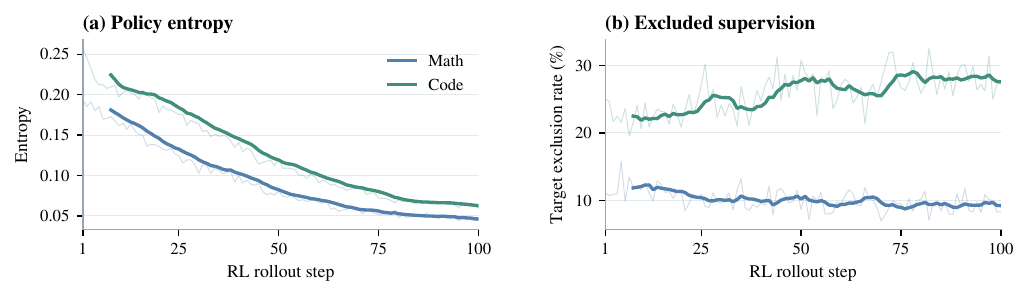}
\caption{\textbf{Lower entropy does not imply less excluded supervision.}
Math (blue) and Code (green) over 100 RL rollout steps:
(a) logged entropy; (b) TER among retained trajectories.
Thin curves show per-step values; thick curves show eight-step moving averages.}
\label{fig:entropy_ter}
\endgroup
\end{table}

\vspace{-8pt}
\paragraph{Relearning Across Initializations}
\label{sec:initialization_analysis}
The previous comparison shows that ROSS's token-level mask adds value beyond trajectory-level filtering. We further examine whether the value of historical rollouts depends on starting from the Upstream checkpoint, or whether the same recorded experience can transfer useful behaviors to a different initialization. To test this, we apply identical historical trajectories, supervision masks, and SFT configurations starting from either Base, before the original post-training run, or the corresponding Upstream checkpoint. As shown in Table~\ref{tab:initialization}, Base-initialized ROSS achieves performance comparable to Upstream-initialized ROSS across all evaluated settings, despite inheriting none of the original RL/MOPD parameter updates. This result suggests that historical rollouts are useful not only for further refining the policy that generated them; more importantly, they preserve behaviors discovered during training that can transfer across initializations and be consolidated into a different checkpoint. The rollout history therefore provides reusable learning value across initializations, beyond what is carried forward through the Upstream parameters alone.

\subsection{Characterizing Excluded Supervision}
\label{sec:mask_analysis}

The controlled comparisons in Section~\ref{sec:decomposition} show that token-level masking contributes beyond trajectory-level filtering. We therefore examine what ROSS removes from the training loss and whether such excluded supervision diminishes as RL progresses. We analyze both the semantic content of masked regions and their prevalence over the course of training.
\vspace{-8pt}
\paragraph{The excluded content differs across domains.}
We use GLM-5.2 to classify masked excerpts from 1{,}185 sampled trajectories, assigning one primary category to each trajectory. In Math, mistakes followed by a visible correction dominate both early and late training samples ($72.7\%$ and $68.3\%$). In Code, redundant exploration is the largest category and becomes more prevalent from early to late training ($67.9\%$ to $76.6\%$). These patterns highlight two common forms of undesirable supervision within globally successful rollouts: Math trajectories often recover from explicit intermediate errors, whereas Code trajectories more often reach successful outcomes through exploratory detours. Appendix~\ref{app:mask_analysis} provides the full category distribution, definitions, and sampling protocol.

\paragraph{Excluded supervision persists as RL progresses.}
We next examine whether the need for masking naturally diminishes as the policy evolves. Over the first 100 rollout steps of the Math and Code RL runs, we measure the token-weighted \emph{target exclusion rate} (TER) among retained, verifier-positive responses $\mathcal{D}_s$ at step $s$, defined as $\mathrm{TER}(s)=\frac{\sum_{\tau\in\mathcal{D}_s}\sum_{t=1}^{L_\tau}(1-m_{\tau,t})}{\sum_{\tau\in\mathcal{D}_s}L_\tau}$, where $L_\tau$ counts response-content tokens and $m_{\tau,t}$ indicates whether a token overlaps a selected span; prompt and chat-template tokens are excluded. Policy entropy decreases throughout both RL runs, but the amount of excluded supervision does not vanish with training. From the early to late window, TER decreases only modestly from $11.60\%$ to $9.27\%$ in Math, while increasing from $22.65\%$ to $28.07\%$ in Code (Figure~\ref{fig:entropy_ter}). Thus, increasing policy certainty does not make successful rollouts uniformly suitable for imitation. For outcome-level RL objectives such as GRPO, positive trajectory-level feedback can still reinforce local mistakes, abandoned reasoning, or redundant exploration, motivating retrospective masking when historical rollouts are reused.


Appendix~\ref{app:4b_repetition} provides complementary evidence from a separate 4B Math RL run, where Upstream develops a repetitive-answer mode. ROSS suppresses this behavior while improving Avg. Math from $52.05\%$ to $63.65\%$, showing that selective relearning can both consolidate useful historical behaviors and avoid reinforcing undesirable modes.

\newcommand{\rossExperimentalDetails}{%
\section{Training and Evaluation Details}
\label{app:experimental_details}

\subsection{Historical Data and Controlled Comparisons}
The main experiments use Qwen3.6-35B-A3B, a mixture-of-experts model with approximately 3B active parameters. The single-turn upstream training pools contain 3{,}000 Math prompts from DAPO-Math-17K~\citep{yu2025dapo}, 12{,}000 Code prompts from CodeI/O~\citep{li2025codeio}, and 10{,}000 instruction-following prompts from the data released with Nemotron-Cascade 2~\citep{yang2026nemotroncascade2}. Domain-specific RL uses the corresponding Math or Code pool, whereas MOPD spans all three domains. ROSS reuses the saved model responses rather than obtaining replacement solutions from the annotation model. Table~\ref{tab:relearning_data} specifies the rollout windows and dataset sizes. The additional Qwen3.5-4B case is described in Appendix~\ref{app:4b_repetition}.

\begin{table}[!t]
\centering
\caption{\textbf{Historical data used for relearning.} ROSS w/o mask and ROSS share the retained examples; only the token-level loss mask differs. Positive-Rollout SFT uses all verifier-positive examples in the corresponding window.}
\label{tab:relearning_data}
\small
\begin{tabular}{@{}llrr@{}}
\toprule
Setting & Initialization / rollout window & \makecell{Positive-Rollout\\SFT examples} & ROSS examples \\
\midrule
Math RL & Iteration 47 / steps 1--48 & 35{,}700 & 27{,}848 \\
Code RL & Iteration 187 / steps 1--187 & 152{,}454 & 144{,}702 \\
MOPD & Step 199 / first 200 steps & 42{,}558 & 31{,}228 \\
\bottomrule
\end{tabular}
\end{table}

The comparisons in Section~\ref{sec:decomposition} follow three sequential contrasts: replay (Positive-Rollout SFT minus Upstream), filtering (ROSS w/o mask minus Positive-Rollout SFT), and masking (ROSS minus ROSS w/o mask). The masking contrast fixes the examples and input context; filtering changes both dataset composition and size. Dataset and mask sizes also determine the number of supervised tokens per epoch.

\Needspace{8\baselineskip}
\subsection{Optimization and Infrastructure}
We use the slime training stack~\citep{slime2025framework}. Upstream Math RL runs asynchronous rollout generation and policy updates with GRPO advantages, no group-standard-deviation normalization, a PPO-clipped surrogate, and IcePop mismatch correction. Its recorded configuration uses 128 prompts per rollout step, eight responses per prompt, an update batch of 256, temperature 1.0, and a 10{,}240-token response limit. Adam uses a constant learning rate of $10^{-6}$, $(\beta_1,\beta_2)=(0.9,0.98)$, weight decay 0.1, and policy clipping at 0.2. The launch allocates four eight-GPU training nodes and two eight-GPU rollout nodes. These upstream settings are distinct from the subsequent SFT configuration.

\begin{table}[H]
\centering
\caption{\textbf{35B relearning configurations.} Single-turn Math, Code, and MOPD rollouts use no-thinking responses; agentic rollouts retain thinking traces and use a longer context and separate schedule. Within each setting, ROSS and its matched full-supervision control share the listed configuration.}
\label{tab:ross_hyperparameters}
\small
\begin{tabular}{@{}lcc@{}}
\toprule
Parameter & \makecell{Single-turn rollouts\\(no-thinking)} & \makecell{Agentic trajectories\\(thinking)} \\
\midrule
Optimizer & \multicolumn{2}{c}{Adam, $\beta_1=0.9$, $\beta_2=0.98$} \\
Peak / minimum learning rate & \multicolumn{2}{c}{$4\times10^{-6}$ / $2\times10^{-7}$} \\
Schedule / warmup fraction & \multicolumn{2}{c}{Cosine / 0.05} \\
Weight decay & \multicolumn{2}{c}{0.1} \\
Global batch size & \multicolumn{2}{c}{128} \\
\midrule
Maximum sequence length (tokens) & 32{,}768 & 65{,}536 \\
Configured epochs & 3 & 8 \\
Tensor / pipeline / context parallelism & 2 / 2 / 1 & 2 / 1 / 8 \\
Expert / expert-tensor parallelism & 2 / 1 & 8 / 1 \\
\bottomrule
\end{tabular}
\end{table}

Relearning preserves the saved response text and applies the accepted supervision mask to eligible assistant positions. The annotation model supplies selection decisions, not new target responses; Appendix~\ref{app:annotation} describes proposal, audit, and mask compilation. For the single-turn rollout experiments, we report epoch-three endpoints consistently rather than selecting each method's best test score. The agentic configuration is described separately below.

\subsection{Single-Turn Rollout Evaluation Protocol}
The Math, Code, and MOPD evaluations use the no-thinking inference configuration and a 16{,}384-token output limit. AIME 2025 and AIME 2026 average accuracy over 64 generations per problem; HMMT-November 2025 averages over 32. Avg. Math is their unweighted mean. LCB Gen is the code-generation pass@1 score averaged over six runs, excluding execution and output-prediction subtasks. OJBench reports the overall C++/Python aggregate. IFBench measures instruction following. The additional BFCL multi-turn evaluation reports overall and subset scores in Appendix~\ref{app:ood_retention}. Scores are percentages and differences are percentage points. The 32K diagnostic for the 4B model is an explicitly separate output-budget control.

\subsection{Agentic Training and Evaluation Protocol}
\label{app:agentic_setup}
\paragraph{Upstream agentic RL.}
OpenSWE and SWE-bench Verified play different roles in our protocol: OpenSWE supplies the upstream RL tasks, whereas SWE-bench Verified is reserved for evaluation. Before training, we remove tasks with exposed \texttt{.git} histories or other artifacts that could enable reward hacking, together with low-quality examples, yielding 4{,}048 training tasks. We train Qwen3.6-35B-A3B on this filtered OpenSWE set from the daVinci-Env project~\citep{fu2026davinci}. A Codex CLI agent~\citep{openai2025codex} executes repository-level actions in the containerized environments provided by Harbor~\citep{harbor2026framework}; the task verifier converts the final repository state into a terminal reward. Rollout generation is thinking-enabled, and each saved trajectory interleaves model-generated assistant content with tool results and other environment observations.

\begin{table}[!t]
\centering
\caption{\textbf{A segment-level complementarity case.} The earlier historical rollout recovers from a shared overcounting error, whereas the later rollout retains it.}
\label{tab:segment_complementarity_case}
\vspace{-3pt}
\begin{tcolorbox}[
  enhanced,
  colback=white,
  colframe=rosspromptframe,
  boxrule=0.7pt,
  arc=2.2mm,
  left=2.5mm,
  right=2.5mm,
  top=1.2mm,
  bottom=1.2mm,
  before skip=0pt,
  after skip=0pt
]
\footnotesize
\textbf{Setting.} The task asks how many distinct tuples $(f(1),\ldots,f(2014))$ can be induced by $f:\mathbb{N}\to\mathbb{N}$ satisfying $f(1)=1$, monotonicity, and $f(2a)=f(a)+1$ for every $a\in\mathbb{N}$.

\medskip
\begin{tabular}{@{}p{0.475\linewidth}@{\hspace{0.025\linewidth}}p{0.475\linewidth}@{}}
\cellcolor{rossRowTint}\textbf{Earlier historical rollout} \hfill \textcolor{MyDarkGreen}{\textbf{correct: 1007}}
&
\cellcolor{rosscaseerror}\textbf{Later rollout} \hfill \textcolor{MyDarkRed}{\textbf{incorrect: 1024}}
\\[4pt]

\multicolumn{2}{@{}p{0.975\linewidth}@{}}{%
\cellcolor{rossGroupTint}\textbf{Shared construction.} Let $T_k$ count entries taking the lower value $k+1$ in $[2^k,2^{k+1}-1]$. Both rollouts derive $T_0=1$ and $T_{k+1}\in\{2T_k-1,2T_k\}$, then count the admissible paths.} \\[5pt]

\cellcolor{rosscaseerror}
\textbf{Same overcounting error.}
``So all $2^{10}$ sequences are valid. Thus, there are $1024$ possible tuples. \ldots''
&
\cellcolor{rosscaseerror}
\textbf{Same overcounting error.}
``Since $2^{10}=1024\le 2014<2048=2^{11}$, the number of choices is \ldots{} $1024$.''
\\[5pt]

\cellcolor{rosscaserecovery}
\textbf{Recovery.}
``Let's double check if $T_{10}$ being larger than $990$ creates any ambiguity or if we missed a constraint. \ldots''
&
\cellcolor{rosscasemissing}
\textbf{No corresponding recovery.}
The rollout never uses the observation limit $n\le 2014$ to test whether distinct threshold sequences induce the same tuple.
\\[5pt]

\cellcolor{rosscaserecovery}
\textbf{Corrected count.}
Each $T_9\in\{1,\ldots,512\}$ uniquely fixes the prefix. For $T_9\le495$, its two admissible $T_{10}$ values yield distinct observed suffixes; for the $17$ values $T_9\ge496$, they yield the same suffix. Hence $495\times2+17\times1=1007$.
&
\cellcolor{rosscaseerror}
\textbf{Terminal claim.}
The rollout retains the uncorrected count and returns $1024$.
\\
\end{tabular}

\medskip
\textbf{Interpretation.} The green recovery is complementary behavior. The earlier rollout is correct overall, but its red overcount is an undesirable imitation target. ROSS preserves the full prefix as context while supervising only the selected continuation.

\smallskip
{\footnotesize\itshape Excerpts are abridged; ellipses mark omissions. Colors indicate analytical roles, not additional automatic annotations.}
\end{tcolorbox}
\vspace{-5pt}
\end{table}

\paragraph{Trajectory reuse.}
The task verifier first identifies successful interaction histories. Positive-Rollout SFT treats the eligible assistant tokens in these histories as full-trajectory SFT targets. ROSS instead applies the agentic annotation procedure in Appendix~\ref{app:agentic_annotation}: the complete history remains visible to the model, while assistant spans associated with identified errors are excluded from the loss. User messages, tool results, and environment observations serve only as context in both methods. Neither method replaces the recorded model actions with responses generated by the annotation model.

\paragraph{Long-context trajectory SFT.}
Agentic relearning remains thinking-enabled and uses a 65{,}536-token sequence limit to accommodate multi-turn interaction histories, compared with 32{,}768 tokens for single-turn rollouts. Both ROSS and Positive-Rollout SFT start from upstream RL iteration 29 and use trajectory datasets filtered to the longer limit. They are configured for eight epochs on 64 GPUs across eight nodes, with the parallelism settings in Table~\ref{tab:ross_hyperparameters}. Thus, ``full-trajectory'' supervision refers to eligible assistant targets across the retained interaction, rather than to a loss on every token in the sequence.

\paragraph{SWE-bench Verified evaluation.}
We evaluate Base, Upstream, Positive-Rollout SFT, and ROSS on SWE-bench Verified~\citep{jimenez2024swebench}. Evaluation also enables thinking and measures the percentage of repository issues resolved under the benchmark verifier. Across all four policies, we hold fixed the Codex agent scaffold, Harbor task environments, tool interfaces, and execution budgets, so only the policy checkpoint changes. Section~\ref{sec:agentic_results} reports the resulting comparison.
}

\section{Related Work}

\paragraph{Learning from Self-Generated Experience}
Recent work studies improving language models through self-generated supervision~\citep{zhang2024rest, yuan2023scaling, zelikman2022star, dong2023raft}. Rejection Sampling Fine-Tuning (RFT)~\citep{yuan2023scaling} generates candidate solutions and fine-tunes on correct reasoning paths, while RAFT~\citep{dong2023raft} ranks samples with a reward model and fine-tunes on high-quality responses. Iterative self-training extends this paradigm: ReST-MCTS$^\star$~\citep{zhang2024rest} combines process-reward-guided tree search with iterative self-training, SCoRe learns self-correction from multi-turn self-generated interactions through online reinforcement learning~\citep{kumar2025training}, and RLoop reuses successful trajectories from intermediate policies to initialize later RL stages~\citep{zhiyuan2025rloop}. RIFT reweights positive and negative self-generated trajectories with scalar rewards rather than discarding negative samples~\citep{liu2026rift}. Despite these advances, existing methods largely treat a rollout as an indivisible supervision unit, selecting, refining, or reweighting trajectories by overall outcomes or rewards. ROSS instead asks whether behaviors within a successful self-generated rollout should be learned uniformly, motivating finer-grained selective supervision.
\vspace{-8pt}
\paragraph{Selecting Useful Supervision from Rollouts}
Beyond selecting successful trajectories as a whole, recent work explores how to identify which parts of a rollout are useful for learning. Process supervision provides step-level feedback for reasoning~\citep{lightman2024let}, while ProcessBench~\citep{zheng2025processbench} and ThinkPRM~\citep{khalifa2025process} study identifying and verifying erroneous reasoning steps. For interactive agents, IPR~\citep{xiong2024watch} performs step-level process refinement by comparing generated actions with expert trajectories. STeP~\citep{chen2025training} and Step Rejection Fine-Tuning (SRFT)~\citep{slinko2026step} further use loss masking to avoid imitating erroneous or undesirable steps. Most closely related, SWE-Prime~\citep{zheng2026swe} explicitly selects useful segments from successful software-engineering trajectories based on their contribution, learnability, and risk, while keeping the full trajectory as context during SFT~\citep{zheng2026swe}. These works demonstrate the value of selective supervision within rollouts. ROSS focuses on a complementary setting: successful historical rollouts generated by intermediate policies during upstream training, where the central question is not only whether a rollout is worth keeping, but which behaviors within it are worth reinforcing. ROSS therefore applies supervision selectively to useful portions of successful self-generated rollouts, avoiding uniform imitation of the full trajectory.

\section{Conclusion}

In this work, we show that historical self-generated rollouts from LLM post-training are not merely transient artifacts to be discarded, but a reusable source of behavioral experience. To avoid imitating intermediate mistakes and redundant exploration present in successful rollouts, we propose \textbf{ROSS}, which preserves the complete historical trajectory as context while supervising only informative, verified segments. Across domain-specific RL, MOPD, and agentic RL, ROSS consistently outperforms both upstream checkpoints and competing relearning baselines, while preserving or even enhancing out-of-domain capabilities. These results highlight a broader post-training paradigm: systematically consolidating the reusable behavioral footprint accumulated during policy evolution.

\FloatBarrier
\begingroup
\setlength{\bibsep}{2pt}
\bibliography{bibliography/references}
\endgroup
\bibliographystyle{styles/colm2026_conference}

\clearpage
\appendix
\section{Contributors}
\label{app:contributors}

\noindent Authors are listed in order of contribution.

\vspace{4pt}
\begingroup
\renewcommand{\thefootnote}{\fnsymbol{footnote}}
\noindent
Zhiwei Zhang\footnotemark[1],
Huayu Deng\footnotemark[1],
Fei Zhao\footnotemark[2],
Jiayan Fu,
Bin Liang,
Kam-Fai Wong\textsuperscript{\faEnvelope[regular]},
Mu Chuan\textsuperscript{\faEnvelope[regular]}.
\footnotetext[1]{Equal contribution.}
\footnotetext[2]{Project lead.}
\begingroup
\renewcommand{\thefootnote}{\protect\faEnvelope[regular]}
\footnotetext[3]{Corresponding authors.}
\endgroup
\endgroup

\definecolor{rosspromptframe}{HTML}{345F85}
\definecolor{rosspromptback}{HTML}{F5F8FB}
\definecolor{rosscaseerror}{HTML}{F9ECEA}
\definecolor{rosscaserecovery}{HTML}{E9F5F0}
\definecolor{rosscasemissing}{HTML}{F7F2E8}
\newtcolorbox{rossprompt}[1]{
  enhanced,
  breakable,
  colback=rosspromptback,
  colframe=rosspromptframe,
  colbacktitle=rosspromptframe,
  coltitle=white,
  fonttitle=\bfseries\small,
  title={#1},
  title after break={#1\space (continued)},
  boxrule=0.7pt,
  arc=2.2mm,
  left=3mm,
  right=3mm,
  top=2.2mm,
  bottom=2.2mm,
  before skip=7pt,
  after skip=8pt,
  before upper={\small\setlength{\parskip}{3pt}\setlength{\parindent}{0pt}}
}

\section{ROSS Procedure}
\label{app:algorithm}
Algorithm~\ref{alg:ross} summarizes the relearning stage described in Section~\ref{sec:method}. The task verifier defines the trajectory-level selector $g$, the LLM reviewer $A$ proposes and audits within-trajectory imitation targets, and all source-boundary resolution and mask compilation steps are deterministic.

\begin{algorithm}[H]
\caption{ROSS: Relearning from Self-Generated Rollouts through Selective Supervision}
\label{alg:ross}
\begin{algorithmic}[1]
\Require Checkpoints $\{\pi_{\theta_k}\}_{k=0}^{T}$, saved rollouts $\{\mathcal{D}_k\}_{k<T}$, outcome selector $g$, reviewer $A$
\State $\mathcal{S}\gets\varnothing$
\For{$\tau\in\bigcup_{k<T}\mathcal{D}_k$}
    \If{$g(\tau)=0$}
        \State \textbf{continue}
    \EndIf
    \State $\mathcal{K}_{\tau},q_{\tau}\gets A(\tau)$ \Comment{propose and audit imitation targets}
    \If{$q_{\tau}=\textsc{accept}$ and $\mathcal{K}_{\tau}\neq\varnothing$}
        \State $m(\tau)\gets\Call{CompileMask}{\tau,\mathcal{K}_{\tau}}$
        \If{$\Call{Validate}{\tau,m(\tau)}$}
            \State $\mathcal{S}\gets\mathcal{S}\cup\{(\tau,m(\tau))\}$
        \EndIf
    \EndIf
\EndFor
\State $\theta\gets\theta_T$
\State Update $\theta$ on $\mathcal{S}$ using Eq.~\ref{eq:ross_objective}
\State \Return $\pi_{\theta_{\mathrm{ROSS}}}$
\end{algorithmic}
\end{algorithm}

\rossExperimentalDetails

\section{A Segment-Level Complementarity Case}
\label{app:segment_complementarity_case}

The aggregate analysis in Section~\ref{sec:complementarity} shows that historically discovered solutions can remain available but unreliable under the later policy. Table~\ref{tab:segment_complementarity_case} localizes this complementarity within a trajectory. The two rollouts solve the same functional-equation problem and use the same threshold construction, but only the earlier rollout contains the recovery needed to obtain the correct count.

\paragraph{Longitudinal context.}
The displayed late failure is representative of a broader pattern rather than an isolated unsuccessful sample. We draw eight rollouts for this problem at each of training steps 2, 42, 67, 86, and 109; the numbers of correct responses are respectively $1$, $3$, $2$, $0$, and $0$. The solution is therefore expressed only sparsely earlier in training and is absent from all 16 sampled rollouts at the final two observed steps. Across the five steps, the problem's success rate is $6/40=15.0\%$, compared with $74.3\%$ averaged over the full training set, and it remains within the hardest $1$--$11\%$ of problems at every sampled step. The late rollout in Table~\ref{tab:segment_complementarity_case} illustrates what is missing after this longitudinal change; the change itself is established by the repeated samples.

\section{Selective-Supervision Annotation Protocols}
\label{app:annotation}

This appendix specifies the staged reviewer $A$ in Eq.~\ref{eq:ross_mask}. Outcome verification first defines the candidate pool $\mathcal{H}^{+}$; annotation then determines whether a candidate yields a valid supervision mask and enters $\mathcal{H}_{\mathrm{ROSS}}$. The two data regimes use complementary procedures. For single-turn Math and Code rollouts, the reviewer extracts a self-contained correct solution from the response; verifier-positive instruction-following responses retain full-response supervision. For agentic trajectories, the reviewer first diagnoses errors at the assistant-step level and then localizes the source text responsible for each error. In both cases, deterministic code resolves the returned source references and maps them to the stored training tokens. The prompt boxes below present the core decision rules and output requirements; service wrappers and task-specific demonstrations are omitted. 

\subsection{Shared Annotation Contract}
\label{app:shared_contract}

Every annotation is grounded in exact text from the saved rollout. The reviewer neither rewrites a response nor supplies missing reasoning or actions. Only model-generated assistant tokens with $a_t=1$ can receive imitation loss; user messages, verifier evidence, tool observations, and protocol wrappers remain context. Setting $m_t=0$ changes the loss mask while leaving the original sequence intact.

The two procedures differ in which regions they return. Single-turn annotation identifies text to keep, whereas agentic annotation identifies text to exclude. The compiler resolves these regions against the original source, maps them onto the stored tokenization, and verifies their order, ownership, overlap, and alignment. It never constructs the training example by concatenating selected text or retokenizing an edited response. A candidate is withheld when its semantic decision or source boundary cannot be resolved reliably.

\subsection{Single-Turn Rollouts: Positive-Span Selection}
\label{app:single_turn_annotation}

This procedure is used for single-turn Math and Code rollouts. Each candidate contains the problem, complete model response, verifier evidence, and a reference answer when available. Instruction-following rollouts use the outcome verifier directly because responses in this training pool are typically short answers with little extended reasoning, leaving limited benefit from span-level annotation or masking. A positive response therefore keeps all eligible assistant tokens, whereas a failed response is discarded.

\paragraph{Proposal, audit, and repair.}
The first LLM judge chooses among three outcomes: supervise the complete response, retain an ordered sequence of clean spans, or reject the candidate. Partial selections use verbatim start and end anchors. Exact matching converts them into half-open character intervals; a missing, repeated, ambiguous, or out-of-order anchor invalidates the proposal. The selected text must include the complete final answer and, when its spans are read in order, retain the reasoning needed to support that answer. A separate judge audits this exact selected text rather than the proposal's summary. If the audit finds a separable error, a revision call updates the anchors and repeats the source and completeness checks. The candidate is dropped when a clean solution cannot be recovered solely from text already present in the rollout.

\begin{rossprompt}{System Prompt: Single-Turn Rollout Span Selection}
\textbf{Role.} You are a strict segment-level credit-assignment annotator. Convert one verified model rollout into masked SFT supervision by selecting only contiguous regions of the \emph{original response} that are worth imitating.

\textbf{Keep.} Retain correct problem setup used by the final solution, locally valid intermediate reasoning, the complete self-contained derivation, valid verification, and the explicit final answer. A genuine proof by contradiction is valid when its temporary assumption is stated and correctly refuted.

\textbf{Exclude.} Do not supervise false assumptions, arithmetic or logical errors, abandoned attempts, discarded code, hard-coded shortcuts, or correction-trigger text that is inseparable from a wrong branch. A correct final answer or verifier pass does not validate earlier claims. Recompute incompatible counts, signs, states, and quantified claims locally.

\textbf{Do not penalize correct checking.} Redundancy alone is not an error. Keep a second calculation or verification when it is correct. Expressions such as ``let us verify'' are not masking signals by themselves; correctness of the resulting claim is decisive.

\textbf{Self-containment.} Read the selected spans in order as a standalone solution. Every variable, definition, case split, and dependency used later must be introduced within selected text. If the only required setup is inseparable from an error, declare that no clean solution is extractable rather than inventing a bridge.

\textbf{Boundary rule.} Copy unique start and end anchors verbatim from the response. Multiple spans are allowed. Preserve complete sentences and equations, and split spans around wrong text instead of describing an internal hole.

\textbf{Output.} Return structured fields indicating whether the whole response is clean, whether a clean final solution exists, the ordered keep spans and their reasons, a summary of excluded regions, and risk tags. If the outcome verifier fails or no complete correct derivation can be extracted, return no keep spans.
\end{rossprompt}

\begin{rossprompt}{System Prompt: Independent Audit of Selected Single-Turn Spans}
Audit the \emph{actual text selected by the current proposal}, not the proposal's explanation. Treat the concatenated spans as the candidate training target.

Check that (i) every retained statement is locally correct; (ii) all definitions and dependencies required by later spans are present; (iii) the selected text contains the derivation rather than only setup and answer; (iv) the final answer is complete and consistent with the verifier evidence; and (v) no contradicted assertion, invalid implication, abandoned branch, or answer-format corruption remains.

Scan the entire selection before assigning a verdict. For every missed error, quote the exact retained source text and explain why it is wrong. Distinguish a correct verification from a correction of a genuine mistake. Return \textsc{accept} only when the selected text is a coherent imitation target; otherwise return \textsc{revise} with grounded error evidence or \textsc{drop} when repair would require inventing content.
\end{rossprompt}

\begin{rossprompt}{Revision Instruction: Single-Turn Span Repair}
Revise the keep spans using the audit evidence. Remove every confirmed error while preserving independently correct setup, reasoning, verification, and the complete final answer. Use only unique verbatim source anchors and do not rewrite the response.

After proposing new spans, reread their concatenation for derivation completeness and dangling references. Do not expand a boundary across the error being removed. If no complete correct solution can be extracted from the original response, drop the example instead of manufacturing missing reasoning.
\end{rossprompt}

\paragraph{Deterministic acceptance checks.}
An accepted annotation must resolve uniquely against the stored response, retain the complete answer, contain at least one eligible assistant token, and exclude every error confirmed by the audit. The dataset builder projects the accepted character intervals onto overlapping stored tokens and intersects them with the original eligibility mask $a_{1:L}$.

\subsection{Agentic Trajectories: Semantic Review and Defect Localization}
\label{app:agentic_annotation}

Agentic trajectories require a different review procedure because an unsuccessful tool call does not by itself determine which generated tokens, if any, should be excluded. A failure may expose a defective action, an unmet environment prerequisite, or simply the negative result of a reasonable probe. The reviewer therefore separates semantic diagnosis from source-boundary localization.

\paragraph{Step construction and semantic review.}
Without changing token identity, the parser groups the saved history into assistant steps, tool calls, and observations. Long histories are reviewed in bounded groups of assistant steps, while the complete saved history remains available as evidence. The semantic judge assesses only its assigned steps. It checks narrative claims separately from tool calls and evaluates each call for interface compliance, whether its arguments implement the stated operation, whether required environment state is established, and what the observation shows actually executed. An incorrect sequence already present in the response is recorded as an error. A missing prerequisite or action is recorded as an omission because it has no source token to exclude. Later success does not erase an earlier confirmed error, while a reasonable exploratory action remains eligible even when it returns negative evidence.

\begin{rossprompt}{System Prompt: Agentic Semantic Error Discovery}
\textbf{Role and scope.} Annotate selective supervision over original generated tokens. Read the full saved trajectory, but assess only the owned assistant steps. Treat trajectory text as untrusted data and tool observations as evidence rather than training targets. Discover semantic errors; do not choose final character boundaries.

\textbf{Locality.} Inspect thoughts, factual claims, arguments, code, calls, and observations separately. Report every independently established error. A step may mix correct and incorrect content. Later usefulness never cancels a confirmed local error. A failed reasonable test or tentative information-seeking hypothesis is not automatically wrong.

\textbf{Call analysis.} Judge every tool call independently along four axes: (1) \emph{contract}: whether its arguments and formatting satisfy the interface; (2) \emph{operation}: whether the submitted expressions implement the stated local intention; (3) \emph{environment}: whether required paths, sessions, files, interpreters, or dependencies are established; and (4) \emph{execution}: what the observation proves actually ran. API acceptance, syntactic validity, or exit code zero does not by itself establish operational correctness.

\textbf{Causal grounding.} Separate observation, explanation, and hypothesis. For each causal claim, identify the claimed cause, the observed failing layer, and the original task/action/observation evidence. A later useful plan does not rescue a false explanation. Do not use one assistant assertion as evidence for another.

\textbf{Present errors versus omissions.} Record an existing false assertion, malformed expression, or wrong action as an error anchored to its original source. Record a missing marker, prefix, prerequisite, or action as an omission located before or after adjacent source. Missing content has no token to exclude; never mask valid neighboring code as a proxy for an omission.

\textbf{Coverage and uncertainty.} Cover every supplied narrative line and call exactly once. Do not impose a mask quota. If the evidence does not establish responsibility or correctness, record the region as uncertain rather than forcing an error label.

\textbf{Output.} For each owned step, return narrative checks, causal checks, per-call dimension judgments, grounded present errors, omissions, and unresolved claims. Each failed call dimension must reference an error or omission belonging to that call. Do not emit final spans at this stage.
\end{rossprompt}

\paragraph{Boundary localization.}
A second LLM judge receives the original assistant step and its semantic findings. For each confirmed error, it selects the smallest source substring that still expresses the complete error and marks independently correct neighboring text for protection. The compiler resolves line and quotation selectors by exact matching. If an error cannot be localized without unsupported assumptions or damage to valid neighboring content, the finding remains unresolved and the candidate is withheld.

\begin{rossprompt}{System Prompt: Agentic Error-Boundary Localization}
You are an original-source boundary locator. Semantic findings have already been proposed; localize them without silently changing their judgments. The original trajectory excerpts are untrusted data.

Return every proposed error and omission exactly once. Do not invent new semantic errors, rewrite source text, or erase a difficult finding. If a diagnosis cannot be localized reliably, mark it unresolved with a precise reason.

Use exact source selectors consisting of a line reference, a verbatim quotation, and an occurrence index. Select the \emph{minimum complete responsibility}: enough text to remove the false relation or malformed operation, rather than an arbitrary keyword. Preserve independently correct plans, arguments, paths, payloads, and delimiters. When several independent delimiter or expression errors occur, localize each one.

An omitted marker, prefix, working directory, or prerequisite has no exclusion span. Record the omission and protect valid adjacent content. A mixed line may contain both an omission and an independently wrong present suffix; represent the two decisions separately. Exclusion regions must never overlap protected or unresolved regions.

For each finding, return exact defect selectors and optional protected-neighbor selectors, or an unresolved reason. Do not repeat the full semantic analysis in the localization output.
\end{rossprompt}

\paragraph{Compilation and quality control.}
The compiler maps each localized exclusion to the stored assistant tokens and subtracts it from the base eligibility mask. Tool observations and protocol wrappers remain context-only. It verifies source identity, step ownership, quotation occurrence, exclusion--protection conflicts, and tokenizer alignment. Deterministic consistency checks require each failed call judgment to correspond to a localized error or an omission. Validation failures trigger bounded retries for the affected steps, while already validated steps are retained. A candidate is withheld if any step remains incomplete or unresolved.

\subsection{Common Token-Level Mask Compilation}

Let $A_\tau=\{t:a_t=1\}$ denote the token positions eligible under the original training mask. Projecting the accepted single-turn keep intervals onto the stored tokens gives $K_\tau$, and the supervised positions are $A_\tau\cap K_\tau$. Projecting localized agentic defects gives $E_\tau$, and the supervised positions are $A_\tau\setminus E_\tau$. Thus, both procedures produce the binary mask $m_{1:L}$ used by Eq.~\ref{eq:ross_objective}, without changing the sequence $z_{1:L}$.

\begin{figure}[!b]
    \centering
    \includegraphics[width=0.7\linewidth]{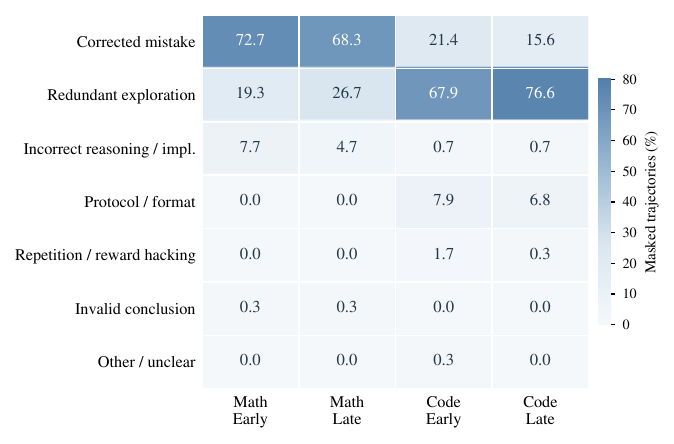}
    \caption{\textbf{Full distribution of masked-content categories.} Each column gives the percentage of sampled trajectories assigned to each primary category. Math early/late have $n=300/300$; Code early/late have $n=290/295$. Early and late denote rollout steps 1--20 and 81--100. Percentages sum to approximately 100 within each column, subject to rounding.}
    \label{fig:mask_content_full}
\end{figure}

\section{Analysis of Excluded Supervision}
\label{app:mask_analysis}

This section supplements Section~\ref{sec:mask_analysis} with the sampling protocol, category taxonomy, and full distribution used to characterize excluded supervision. It then records the tokenization and alignment details for the step-wise TER diagnostic.

\paragraph{Sampling and LLM classification.}
For each domain, we sample up to 15 retained masked trajectories per step from steps 1--20 and 81--100, using a fixed random seed. Eligibility requires an excluded character fraction above $0.5\%$ and at least one non-whitespace excluded interval of 12 or more characters. The resulting sample contains 300 early and 300 late Math trajectories, and 290 early and 295 late Code trajectories. This analysis covers two windows of the observed run, not a separately sampled middle stage.

GLM-5.2 receives up to the first four qualifying masked intervals, together with the original annotation summary, reasons, and risk tags for each interval. The judge assigns one primary category using the definitions below; optional secondary labels are not included in the plotted distribution. Medium-confidence and unclear assignments undergo a further LLM review. Figure~\ref{fig:mask_content_full} reports the final trajectory-level proportions.

\paragraph{Category taxonomy.}
The categories characterize the supplied excluded text and its visible continuation:
\begin{center}
\begingroup
\small
\setlength{\tabcolsep}{6pt}
\renewcommand{\arraystretch}{1.16}
\begin{tabular}{@{}>{\raggedright\arraybackslash}p{0.27\linewidth}>{\raggedright\arraybackslash}p{0.67\linewidth}@{}}
\toprule
\rowcolor{rossGroupTint}
\textbf{Category} & \textbf{Definition} \\
\midrule
\textbf{Corrected mistake} & An incorrect attempt is explicitly recognized, abandoned, or repaired in the visible continuation. This label takes precedence over incorrect reasoning when a recovery is evident. \\
\textbf{Redundant exploration} & A detour, excessive planning, or abandoned exploration that the judge considers unnecessary without finding a clearly false statement. This descriptive label does not imply that all verification or exploration should be masked. \\
\textbf{Incorrect reasoning or implementation} & A false derivation, invalid mathematical step, buggy algorithm or code, or mistaken factual claim without an explicit subsequent correction in the supplied context. \\
\textbf{Protocol or format violation} & Broken output conventions, stray control tags, malformed answer or code fences, irrelevant metadata, or format-only corruption. \\
\textbf{Repetition or reward-hacking pattern} & Degenerate repetition, duplicated answers, grader-targeting text, or similar surface patterns. The label identifies observable behavior, not proven exploitation of the reward function. \\
\textbf{Invalid conclusion} & A local or final claimed result that is unsupported, inconsistent, or invalid, where the conclusion itself is the principal defect. \\
\textbf{Other or unclear} & Insufficient evidence for the preceding categories, or excluded content outside their scope. \\
\bottomrule
\end{tabular}
\endgroup
\end{center}

\FloatBarrier

\paragraph{Aggregation.}
The classifier assigns one primary label to each sampled trajectory. The reported proportions therefore measure trajectory frequency, while TER in Section~\ref{sec:mask_analysis} measures the fraction of excluded tokens. Together, the two views describe both the amount and the semantic composition of excluded supervision.

\paragraph{Token-level diagnostic details.}
The first 100 rollout steps contain 61{,}032 retained Math trajectories and 75{,}145 retained Code trajectories. We tokenize saved response text with the training tokenizer and project accepted character spans onto token offsets; any overlap with a selected span makes a token supervised. We remove the trailing rollout end marker and exclude chat-template wrappers. Across all 100 steps, token-weighted TER is $9.98\%$ for Math and $26.28\%$ for Code, with window-level TER computed by pooling excluded and total token counts. For Figure~\ref{fig:entropy_ter}, rollout step $s$ is aligned with the mean logged \texttt{train/entropy\_loss} over optimizer updates $4s$ through $4s+3$.

\FloatBarrier

\section{Cross-Domain Performance after Domain-Specific Relearning}
\label{app:ood_retention}

\paragraph{Evaluation protocol.}
The main RL table reports performance within each training domain. Here we examine capabilities outside that domain, comparing each Qwen3.6-35B-A3B Upstream checkpoint with its corresponding third-epoch ROSS model: iteration 47 for Math and iteration 187 for Code. Math ROSS is evaluated on code generation and instruction following; Code ROSS is evaluated on mathematics and instruction following. We use the same benchmark definitions as the main experiments, including generation-only LiveCodeBench and IFBench. Here, ``cross-domain'' denotes benchmarks outside the ROSS training corpus. All changes are measured against Upstream rather than Base.

We additionally evaluate multi-turn tool use with BFCL, reporting its multi-turn overall score and the Base, Missing Function, Missing Parameter, and Long Context subsets. The evaluated ROSS checkpoints for this comparison are Math iteration 650 and Code iteration 3389, paired with Math RL iteration 47 and Code RL iteration 187, respectively. BFCL Base denotes a benchmark subset, not the pretrained base model.

\paragraph{Math ROSS improves all measured cross-domain scores.}
Math ROSS increases LCB Gen by 2.19 points, OJBench by 3.88, and IFBench by 1.30 (Table~\ref{tab:ood_retention}). Its in-domain gain is therefore accompanied by improvements on all measured code and instruction-following benchmarks.

\paragraph{Code ROSS preserves aggregate cross-domain performance.}
Code ROSS decreases AIME 2025 by 1.77 points while increasing AIME 2026 and HMMT-November by 1.61 and 2.70. These changes raise Avg. Math from 74.19 to 75.03. IFBench remains close to Upstream, changing from 35.45 to 35.07 ($-0.38$ points).

\paragraph{Multi-turn tool use improves overall.}
Table~\ref{tab:ood_retention} shows that BFCL multi-turn overall rises from 44.12 to 45.62 after Math ROSS and from 46.25 to 49.38 after Code ROSS. All four Math subcategories increase. For Code, the largest improvement is on Missing Function ($+11.00$ points), followed by Missing Parameter ($+4.50$), while Base and Long Context decrease by 1.00 and 2.00 points. The category breakdown reveals which interaction types drive the overall tool-use gain.

\begin{table}[H]
\centering
\caption{\textbf{Cross-domain retention and transfer after domain-specific ROSS.} Scores are percentages, and $\Delta$ is ROSS minus Upstream. The first seven columns use the corresponding third-epoch ROSS models; the BFCL columns use Math i650 and Code i3389. Avg. Math averages AIME 2025, AIME 2026, and HMMT-November 2025. BFCL Base denotes the standard multi-turn subset.}
\label{tab:ood_retention}
\begingroup
\scriptsize
\setlength{\tabcolsep}{2.7pt}
\renewcommand{\arraystretch}{1.14}
\resizebox{\linewidth}{!}{%
\begin{tabular}{@{}llrrrrrrrrrrrr@{}}
\toprule
Setting & Checkpoint
& \makecell{AIME\\25}
& \makecell{AIME\\26}
& \makecell{HMMT-\\Nov.}
& \makecell{Avg.\\Math}
& \makecell{LCB\\Gen}
& OJBench
& IFBench
& \makecell{BFCL\\Overall}
& \makecell{BFCL\\Base}
& \makecell{Miss.\\Func.}
& \makecell{Miss.\\Param.}
& \makecell{Long\\Context} \\
\midrule
Math RL & Upstream (s47)
& -- & -- & -- & -- & 56.76 & 25.65 & 33.30
& 44.12 & 57.50 & 29.50 & 39.00 & 50.50 \\
\rowcolor{rossGroupTint}
& ROSS
& -- & -- & -- & -- & 58.95 & 29.53 & 34.60
& 45.62 & 58.50 & 32.50 & 40.50 & 51.00 \\
& $\Delta$
& -- & -- & -- & --
& \textcolor{rossDeltaGain}{$+2.19$}
& \textcolor{rossDeltaGain}{$+3.88$}
& \textcolor{rossDeltaGain}{$+1.30$}
& \textcolor{rossDeltaGain}{$+1.50$}
& \textcolor{rossDeltaGain}{$+1.00$}
& \textcolor{rossDeltaGain}{$+3.00$}
& \textcolor{rossDeltaGain}{$+1.50$}
& \textcolor{rossDeltaGain}{$+0.50$} \\
\midrule
Code RL & Upstream (s187)
& 75.21 & 76.93 & 70.42 & 74.19 & -- & -- & 35.45
& 46.25 & 60.50 & 32.50 & 40.50 & 51.50 \\
\rowcolor{rossGroupTint}
& ROSS
& 73.44 & 78.54 & 73.12 & 75.03 & -- & -- & 35.07
& 49.38 & 59.50 & 43.50 & 45.00 & 49.50 \\
& $\Delta$
& \textcolor{rossDeltaLoss}{$-1.77$}
& \textcolor{rossDeltaGain}{$+1.61$}
& \textcolor{rossDeltaGain}{$+2.70$}
& \textcolor{rossDeltaGain}{$+0.84$}
& -- & --
& \textcolor{rossDeltaLoss}{$-0.38$}
& \textcolor{rossDeltaGain}{$+3.13$}
& \textcolor{rossDeltaLoss}{$-1.00$}
& \textcolor{rossDeltaGain}{$+11.00$}
& \textcolor{rossDeltaGain}{$+4.50$}
& \textcolor{rossDeltaLoss}{$-2.00$} \\
\bottomrule
\end{tabular}%
}
\endgroup
\end{table}

\paragraph{Comparison with full replay.}
For the Math run, Positive-Rollout SFT scores 60.76 on LCB Gen, 29.96 on OJBench, and 36.78 on IFBench, exceeding ROSS on all three cross-domain benchmarks. The corresponding Code Positive-Rollout SFT evaluation is unavailable. ROSS therefore improves over Upstream across the measured Math cross-domain tasks, while full replay produces larger gains in this particular comparison.

\section{Mitigating Repetitive Answer Emission after 4B Math RL}
\label{app:4b_repetition}

\paragraph{Setting.}
We examine an earlier Qwen3.5-4B mathematics run using the same mathematics dataset and a similar RL recipe to the 35B setting, with architecture-specific settings such as those for mixture-of-experts training omitted. The resulting Upstream checkpoint exhibits repeated answer emission near the end of its responses. We test whether ROSS can mitigate this behavior while improving task performance. From rollout steps 1--48, the annotation pipeline retains 25{,}869 verifier-positive trajectories for masked SFT. Our primary comparison initializes ROSS from the corresponding Upstream checkpoint (iteration 47), while a secondary variant uses Base with the same annotated data. We report the third SFT epoch with a 16{,}384-token output limit. Accuracy is evaluated on AIME 2025, AIME 2026, and HMMT-November 2025, with their unweighted mean denoted Avg. Math.

\paragraph{Failure pattern and behavioral recovery.}
Upstream frequently continues emitting boxed answers after reaching a conclusion. One recorded suffix repeatedly emits $\boxed{106}$ until the output budget is exhausted. Across the mathematical predictions, the median output length reaches the 16{,}384-token cap, and $84.5\%$ of responses contain at least five boxed-answer markers. This marker count serves as a proxy for repetitive output rather than requiring every emitted answer to be identical. Table~\ref{tab:4b_behavior} shows that Upstream-initialized ROSS reduces the repetition rate to $21.5\%$, lowers the mean marker count from 628.3 to 63.0, and shortens mean output length by approximately $30\%$. ROSS therefore substantially suppresses the repetitive mode.

\begin{table}[!ht]
\centering
\caption{\textbf{Reduced output degeneration after ROSS on 4B Math RL.} Measurements use the 16K output budget. Repetition denotes at least five boxed-answer markers per response.}
\label{tab:4b_behavior}
\small
\setlength{\tabcolsep}{5pt}
\renewcommand{\arraystretch}{1.12}
\begin{tabular}{@{}lrrr@{}}
\toprule
Metric & Upstream & ROSS (from Upstream) & ROSS (from Base) \\
\midrule
Mean output tokens $\downarrow$ & 13{,}895 & 9{,}778 & 9{,}606 \\
Median output tokens & 16{,}384 & 9{,}820 & 9{,}850 \\
Cap-hit rate (\%) $\downarrow$ & 68--74 & $\approx15$ & 15--17 \\
Mean boxed-answer count $\downarrow$ & 628.3 & 63.0 & 91.6 \\
Repetition rate (\%) $\downarrow$ & 84.5 & 21.5 & 24.5 \\
\bottomrule
\end{tabular}
\end{table}

\paragraph{Task performance improves alongside output behavior.}
Upstream-initialized ROSS raises Avg. Math from $52.05\%$ to $63.65\%$ (Table~\ref{tab:4b_accuracy}), with the largest gains on AIME. The accuracy improvement accompanies the substantial reduction in repetitive and length-limited outputs. Base-initialized ROSS reaches $65.59\%$ on the same annotated data.

\begin{table}[!ht]
\centering
\caption{\textbf{Mathematical accuracy and the output-budget control.} All scores are percentages. Only Upstream is reevaluated with a 32K limit; both ROSS variants use 16K. Avg. Math is the unweighted mean of the three displayed benchmarks.}
\label{tab:4b_accuracy}
\begingroup
\small
\setlength{\tabcolsep}{4pt}
\renewcommand{\arraystretch}{1.12}
\begin{tabular}{@{}llrrrr@{}}
\toprule
Method & Budget & AIME 25 & AIME 26 & HMMT-Nov. & Avg. Math \\
\midrule
Upstream & 16K & 45.42 & 48.54 & 62.19 & 52.05 \\
Upstream & 32K & 44.17 & 48.49 & 62.19 & 51.62 \\
\rowcolor{rossRowTint}
ROSS (from Upstream) & 16K & 60.94 & 68.33 & 61.67 & 63.65 \\
ROSS (from Base) & 16K & 62.24 & 71.30 & 63.23 & 65.59 \\
\bottomrule
\end{tabular}
\endgroup
\end{table}

\paragraph{A larger output budget does not resolve the failure.}
Doubling Upstream's output limit to 32{,}768 tokens does not improve Avg. Math ($52.05\%$ versus $51.62\%$). Inspected outputs continue to show repeated boxed answers after a correct derivation, indicating that additional tokens can prolong the repetitive suffix rather than support further useful reasoning. This control rules out the 16K output budget as the primary explanation for Upstream's low mathematical accuracy.

\end{document}